\documentclass{article}
\usepackage{kr}

\usepackage{times}
\usepackage{soul}
\usepackage{url}
\usepackage[hidelinks]{hyperref}
\usepackage[utf8]{inputenc}
\usepackage[small]{caption}
\usepackage{graphicx}
\usepackage{amsmath}
\usepackage{amsthm}
\usepackage{booktabs}
\usepackage{algorithm}
\usepackage{algorithmic}
\usepackage[normalem]{ulem}
\usepackage{amsthm}
\theoremstyle{definition}
\newtheorem{definition}{Definition}
\newtheorem{example}{Example}
\newtheorem{proposition}{Proposition}
\newtheorem{theorem}{Theorem}
\newtheorem{notation}{Notation}

\usepackage{amssymb}
\usepackage{amsmath}
\usepackage{relsize}
\usepackage{mathtools}
\usepackage{enumerate}

\newtheorem*{theorem*}{Theorem}
\newtheorem*{definition*}{Definition}
\newtheorem*{notation*}{Notation}

\usepackage{multicol}

\usepackage{xargs} 
\usepackage[pdftex,dvipsnames]{xcolor} 
\usepackage[colorinlistoftodos]{todonotes}
\setuptodonotes{size=tiny}

\newcommandx{\complete}[2][1=]{\todo[linecolor=orange,backgroundcolor=orange!25,bordercolor=orange,#1]{Complete: #2}}
\newcommandx{\tocite}[2][1=]{\todo[linecolor=pink,backgroundcolor=pink!25,bordercolor=pink,#1]{Cite: #2}}
\newcommandx{\unsure}[2][1=]{\todo[linecolor=blue,backgroundcolor=blue!5,bordercolor=blue,#1]{Unsure: #2}}
\newcommandx{\change}[2][1=]{\todo[linecolor=red,backgroundcolor=red!25,bordercolor=red,#1]{Change: #2}}
\newcommandx{\info}[2][1=]{\todo[linecolor=OliveGreen,backgroundcolor=OliveGreen!25,bordercolor=OliveGreen,#1]{Info: #2}}
\newcommandx{\improvement}[2][1=]{\todo[linecolor=Plum,backgroundcolor=Plum!25,bordercolor=Plum,#1]{Improve: #2}}
\newcommandx{\cut}[2][1=]{\todo[linecolor=yellow,backgroundcolor=yellow!25,bordercolor=yellow,#1]{Potential Cut: #2}}

\usepackage{tikz}
\usetikzlibrary{decorations.pathmorphing,shapes,positioning,arrows.meta,arrows,automata,positioning,calc}

\usepackage{caption}
\usepackage{subcaption}

\newcommand{\bigeq}[2]{=_{[#1:#2]}}
\newcommand{\bigsucc}[2]{\succ_{[#1:#2]}}
\newcommand{\bigsucccurlyeq}[2]{\succcurlyeq_{[#1:#2]}}

\newcommand{\attacks}[1][]{\rightsquigarrow_{#1}}

\newcommand{\labattacks}[2]{\stackrel{\mathclap{\normalfont\mbox{\tiny #1}}}{\rightsquigarrow_{#2}}}

\newcommand{\potentialattacks}[1]{\labattacks{p}{#1}}
\newcommand{\incoherentattacks}{\labattacks{I}{\empty}}
\newcommand{\newattacks}{\labattacks{N}{\empty}}

\newcommand{\x}[1]{x_{#1}}

\newcommand{\xalpha}{\x{\alpha}}
\newcommand{\xbeta}{\x{\beta}}
\newcommand{\xgamma}{\x{\gamma}}

\newcommand{\case}[2]{(\x{#1}, y_{#2})}
\newcommand{\casealpha}{\case{\alpha}{\alpha}}
\newcommand{\casebeta}{\case{\beta}{\beta}}
\newcommand{\casegamma}{\case{\gamma}{\alpha}}

\newcommand{\hlx}[1]{(H_{#1}, L_{#1})}
\newcommand{\hlxalpha}{\hlx{\alpha}}
\newcommand{\hlxbeta}{\hlx{\beta}}

\newcommand{\casedefault}{(x_{\delta}, \delta)}
\newcommand{\casenew}{\case{N}{?}}

\newcommand{\dynamiccase}[2]{(F_{#1}, S_{#1}, y_{#2})}
\newcommand{\dynamiccasealpha}{\dynamiccase{\alpha}{\alpha}}
\newcommand{\dynamiccasebeta}{\dynamiccase{\beta}{\beta}}
\newcommand{\dynamiccasegamma}{\dynamiccase{\gamma}{\alpha}}

\newcommand{\dynamiccasenew}{\dynamiccase{N}{?}}
\newcommand{\dynamiccasedefault}{(\emptyset, \langle \rangle, \delta)}

\newcommand{\af}[2][\empty]{$AF_{#1}(D, \x{#2})$}
\newcommand{\afs}[1]{$AF_{\mathbb{S}}(D, (F_{#1}, S_{#1}))$}

\newcommand{\faacbr}[1]{\mbox{AA-CBR$(D, \x{#1})$}}
\newcommand{\aacbrp}{\mbox{AA-CBR-$\pseq$}}
\newcommand{\faacbrp}[1]{\mbox{AA-CBR-$\pseq(D, \x{#1})$}}

\newcommand{\raacbrp}[1][\empty]{\mbox{AA-CBR-$\pseq_{\langle #1 \rangle}$}}
\newcommand{\rfaacbrp}[2][\empty]{\mbox{AA-CBR-$\pseq_{\langle #1 \rangle}(D, \x{#2})$}}

\newcommand{\afp}[2][\empty]{\af[\pseq]{#2}}
\newcommand{\rafp}[2][\empty]{\af[\pseq{\langle #1 \rangle}]{#2}}

\newcommand{\pseq}{\mathcal{P}}

\newcommand{\adamnew}[1]{#1}
\newcommand{\FT}[1]{#1}

\newtheorem*{theoremman*}{Theorem}
\newtheorem*{definitionman*}{Definition}
\newtheorem*{notationman*}{Notation}

\title{Preference-Based Abstract Argumentation for Case-Based Reasoning \\(with Appendix)}

\author{%
Adam Gould$^1$ \and
Guilherme Paulino-Passos$^1$ \and
Seema Dadhania$^2$\and
\\
Matthew 
Williams$^2$ \and
Francesca Toni$^1$
\affiliations
$^1$Department of Computing, Imperial College London, UK\\
$^2$Department of Surgery and Cancer, Imperial College London, UK\\
\emails
\{adam.gould19, gppassos, s.dadhania, matthew.williams, f.toni \}@imperial.ac.uk,
}

\begin{document}

\maketitle

\begin{abstract}


    In the pursuit of enhancing the efficacy and flexibility of interpretable, data-driven classification models, this work introduces a novel incorporation of user-defined preferences with \emph{Abstract Argumentation} and \emph{Case-Based Reasoning} (CBR). Specifically, we introduce \emph{Preference-Based Abstract Argumentation for Case-Based Reasoning} (which we call \aacbrp), allowing users to define multiple approaches to compare cases with an ordering that specifies their preference over these comparison approaches. We prove that the model inherently follows these preferences when making predictions and show that previous abstract argumentation for case-based reasoning approaches are insufficient at expressing preferences over constituents of an argument. We then demonstrate how this can be applied to a real-world medical dataset sourced from a clinical trial evaluating differing assessment methods of patients with a primary brain tumour. We show empirically that our approach outperforms other interpretable machine learning models on this dataset. 







\end{abstract}

\section{Introduction}
\label{section:introduction}







\emph{Abstract argumentation} is a formalism for representing arguments and 
relationships between them, and for computing which arguments to accept~\cite{DUNG-aa}. It has been shown to be effective 
for recommendation systems~\cite{argument-recommendations}, decision-making tasks~\cite{argument-decision-making}, reasoning with incomplete knowledge~\cite{argument-incomplete-knowledge-recommender} and approaches to explainable artificial intelligence ~\cite{argumentative-xai-survey}. 

\emph{Case-based reasoning} (CBR) is a problem-solving methodology where new problems are solved by retrieving and adapting solutions from similar past cases. Approaches to combine CBR with abstract argumentation are successful for explaining the output of machine learning 
\cite{Prakken2022ATM} or for making predictions 
\cite{DEAr}. \emph{Abstract Argumentation for Case-Based Reasoning} (AA-CBR)~\cite{aa-cbr} is a data-driven interpretable classification and explanation model which has been shown to have utility in many tasks, for example, as an interpretable binary classifier~\cite{ANNA,DEAr,learning-aa-cbr-dt}, for cautiously monotonic reasoning
~\cite{monotonicity-and-noise-tolerance} or for explaining known legislative outcomes~\cite{arbitrated-argumentative-dispute}. 

As a result, AA-CBR represents a novel paradigm as an intrinsically explainable classification model. The need for such models in high-stakes decision-making is becoming increasingly apparent. In medical domains, for instance, decision-support tools need to be understood by a clinician for trust to be established and disagreements to be resolved~\cite{explainability-ai-in-healthcare}. Moreover, legal requirements, such as those established by GDPR~\cite{gdpr-right-to-exp}, require explainable AI to varying extents. 

However, previous approaches to AA-CBR are missing the ability to compare cases by user-defined preferences. Preferences allow stakeholders to influence a decision-making tool and inject domain-specific knowledge, resulting in more desirable reasoning systems or better-performing models. Preferences have been integrated with argumentation systems through a variety of approaches, for example, preferences defined over abstract arguments in preference-based argumentation frameworks~\cite{preference-argumentation-semantics}; preferences defined over structured arguments in \mbox{ASPIC+}~\cite{ASPIC+}; preferences defined over values assigned to arguments as in value-based argumentation frameworks~\cite{value-based-argumentation}; preferences defined by defeats of attack relations as with extended argumentation frameworks~\cite{preference-af-by-attacks}; preferences defined over constituents of an argument such as with Assumption-Based Argumentation with Preferences (ABA+)~\cite{aba+}; and preferences over constituents of cases in which arguments are derived such as with case models\footnote{Literature on AA-CBR, including this work, uses the terms `arguments' and `cases' interchangeably, however, in work on case models, these concepts are separate.}~\cite{case-models-precedent-models}.


    


Within healthcare, preferences allow for specialist expertise and patient goals to affect the decision-making process, leading to more agreeable courses of action~\cite{patient-preferences}. Moreover, preferences with argumentation have many applications within the medical domain~\cite{aba+-patients,Gorgias,aggregating-evidence,lung-cancer-arguing-preferences}. 

Despite this, there is yet to be an AA-CBR-based approach that can integrate preferences over the constituents of arguments. Considering the benefits of AA-CBR as an interpretable classification model, adding preferences will improve the model's utility, flexibility, and performance. In this work, we address the limitations of AA-CBR in this space and contribute the following:

\begin{itemize}
    \item Introduction of \emph{Preference-Based Abstract Argumentation for Case-Based Reasoning} (\aacbrp), a model that allows preferences to be defined over comparison methods of cases. 
    \item Demonstration that \aacbrp~cannot be captured by existing AA-CBR methods. 
    \item Prove that \aacbrp~inherently respects preferences when making classifications, where possible. 
    \item Conduct an empirical study on a real-world medical data set sourced from a clinical trial on patients with a primary brain tumour. 
    \item Showcase how \aacbrp~models with preferences improve performance compared to existing AA-CBR approaches and outperform other interpretable machine learning models. 
\end{itemize}






In Section~\ref{section:preliminaries} we provide the necessary preliminaries. We then demonstrate the motivation for integrating preferences with AA-CBR in Section~\ref{section:motivation}. In Section~\ref{section:methodology}, we introduce~\aacbrp. We explore properties in Section~\ref{section:properties} and conduct an empirical evaluation of the models on a real-world medical dataset in Section~\ref{section:empirical}. Section~\ref{section:related-works} discusses related work. We conclude in Section~\ref{section:conclusion}.

\section{Preliminaries}
\label{section:preliminaries}



\subsection{Abstract Argumentation
}

An \emph{abstract argumentation framework (AF)}~\cite{DUNG-aa} is a pair $\langle Args, \rightsquigarrow \rangle$ where $Args$ is a set of arguments and $\rightsquigarrow \subseteq Args \times Args$ is a binary relation between arguments. For arguments, $\alpha, \beta \in Args$, $\alpha$ \emph{attacks} $\beta$ if $\alpha \rightsquigarrow \beta$. An AF can be represented as a directed graph where nodes are arguments and edges represent attacks. A set of arguments $E \subseteq Args$ \emph{defends} an argument $\beta \in Args$ if for all $\alpha \rightsquigarrow \beta$ there exists $\gamma \in E$ such that $\gamma \rightsquigarrow \alpha$. For determining which arguments to accept, we focus on the \emph{grounded extension}~\cite{DUNG-aa}, which can be iteratively computed as $\mathbb{G} = \bigcup_{i \ge 0} G_i$, where $G_0$ is the set of unattacked arguments and $\forall i \ge 0$, $G_{i+1}$ is the set of all arguments that $G_i$ defends. 




\subsection{Abstract Argumentation for Case-Based Reasoning (AA-CBR)}

AA-CBR is a binary classification model utilising argumentation as the reasoner. It operates on a \emph{casebase} $D$ comprised of labelled examples of a generic \emph{characterisation}. Given $D$ and a new \emph{unlabelled example}, $N$, AA-CBR assigns a label to the new case. Formally:


\begin{definition}[Adapted from~\cite{monotonicity-and-noise-tolerance,DEAr}]
    \label{def:aa-cbr-prediction}
    Let $D \subseteq X \times Y$ be a finite \emph{casebase} of labelled examples where $X$ is a set of \emph{characterisations} and $Y = \{\delta, \bar{\delta}\}$ is the set of possible outcomes. Each example is of the form $(x, y)$. Let $\casedefault$ be the \emph{default argument} with $\delta$ the \emph{default outcome}. Let $N$ be an \emph{unlabelled example} of the form $\casenew$ with $y_{?}$ an unknown outcome. The function AA-CBR$(D, \x{N})$ assigns the new case an outcome as follows: 
    \[ 
        \text{AA-CBR}(D, \x{N}) =
        \begin{cases}
            \delta & \text{if } \casedefault \in \mathbb{G}, \\
            \bar \delta & \text{otherwise.}
        \end{cases}
    \]
    where $\mathbb{G}$ is the grounded extension of the argumentation framework derived from $D$ and $\x{N}$, known as \af{N}. 
\end{definition}

\af{N} is constructed with cases of differing outcomes modelled as arguments. It is assumed that characterisations of the data points are equipped with a partial order~$\succcurlyeq$, which determines the direction of attacks within the casebase \adamnew{and is used to ensure attacks occur between cases with minimal difference}. The new case $\casenew$ is added to the AF by attacking \adamnew{cases considered \emph{irrelevant} to it, as defined by a provided irrelevance relation~$\nsim$.}   


\begin{definition}[Adapted from~\cite{monotonicity-and-noise-tolerance,DEAr}] 
    Let $\succcurlyeq$ and $\nsim$ be a partial order and binary relation defined over $X$, respectively. The argumentation framework \af{N} mined from $D$ and $x_N$ is $(Args, \rightsquigarrow)$ in which:
    \label{def:aa-cbr_geq}

    \begin{itemize}
        \item $Args = D \cup \{\casedefault\} \cup \{N\}$
        \item for $\casealpha, \casebeta \in D \cup \{(x_{\delta}, \delta)\}$, it holds that $\casealpha \rightsquigarrow \casebeta$ iff
              \begin{enumerate}
                  \item $y_{\alpha} \not = y_{\beta}$, and
                  \item either $\x{\alpha} \succ \xbeta$ and $\not\exists \casegamma \in D \cup \{(\x{\delta}, \delta)\}$ with $\xalpha \succ \xgamma \succ \xbeta$ \label{def:concision} \hfill
                  \item or $\xalpha = \xbeta$;
              \end{enumerate}
        \item for $\casealpha \in D \cup \{(\x{\delta}, {\delta})\}$, it holds that $N \rightsquigarrow \casealpha$ iff $N \nsim \casealpha.$
    \end{itemize}

\end{definition}

A casebase $D$ is \emph{coherent} iff there are no two cases $\casealpha, \casebeta \in D$ such that $\xalpha = \xbeta$ and $y_{\alpha} \neq y_{\beta}$, and it is \emph{incoherent} otherwise. 

\adamnew{The second bullet of Definition~\ref{def:aa-cbr_geq} defines attacks between the cases in the casebase. Condition \textbf{2} ensures that attacks occur from greater cases to smaller ones according to the provided partial order. When multiple possible attacks occur, we enforce that attacks originate from the case with minimal difference compared to the attacked case. We refer to this as the most concise possible attack. Condition \textbf{3} defines symmetric attacks for an incoherent casebase.}

\subsection{AA-CBR with Stages}

An extension to AA-CBR~\cite{arbitrated-argumentative-dispute} employs \textit{stages} in the characterisations to represent dynamic features. Stages represent the time at which a case was recorded. If two cases have the same set of features, then the time measure is used to distinguish the cases. \adamnew{If the only difference between two cases is the time measure and the outcome, then it is possible that the change in outcome by the latter case is a result of features that have not been recorded. This approach recognises that not all data may be available and so uses a time measure as a proxy to reason about unknown data.} 
\adamnew{This version of AA-CBR is not defined generally for any characterisation but instead only for a set of features and stages.}
 
Specifically, each case is characterised by  $\mathbb{F} \times \mathbb{S}$, where \adamnew{$\mathbb{F}$ is a set of features and} $\mathbb{S}$ represents the set of all subsequences of $\langle s_{1}, \ldots, s_{n} \rangle$. A subsequence refers to either the empty sequence $\langle \rangle$ or a contiguous sequence of elements, starting from $s_{1}$ and can contain any number of consecutive elements up to and including $s_n$. A case 
may 
be represented as $((F_{\alpha}, S_{\alpha}), y_{\alpha})$, where $F_{\alpha}$ is the set of features, $S_{\alpha}$ is the stages and $y_{\alpha}$ is the outcome. For clarity, we will represent cases without the inner brackets, i.e.~$(F_{\alpha}, S_{\alpha}, y_{\alpha})$. The subsequence relation $\sqsubseteq$ can be defined as in~\cite{arbitrated-argumentative-dispute}: for $S_{\alpha}, S_{\beta} \in \mathbb{S}$, $S_{\beta} \sqsubseteq S_{\alpha}$ iff either $S_{\beta} = \langle s_{1}, \ldots, s_{k} \rangle$, $S_{\alpha} = \langle s_{1}, \ldots, s_{m} \rangle$ and $k \leq m \leq n$ or $S_{\beta} = \langle \rangle$, the empty sequence. We can then define the \afs{N} mined from $D$ and $N$ for AA-CBR with Stages, as follows:

\begin{definition}[Adapted from~\cite{arbitrated-argumentative-dispute}] 

\label{def:aa-cbr-with-stages}
The argumentation framework \afs{N} 
corresponding to a casebase $D$, a default outcome $\delta \in Y$ and a new case $N = \dynamiccasenew$, is $(Args, \rightsquigarrow)$ 
such that:

\begin{itemize}
    \item $Args = D \cup \{\dynamiccasedefault\} \cup \{\dynamiccasenew\}$
    \item $(\emptyset, \langle \rangle, \delta)$ is called the default argument
    \item For $\dynamiccasealpha, \dynamiccasebeta \in Args$, it holds that $\dynamiccasealpha \rightsquigarrow \dynamiccasebeta$ iff
        \begin{enumerate}
            \item $y_{\alpha} \not= y_{\beta}$, and \hspace*{\fill} (different outcomes) 
            \item either
            \begin{enumerate}
                \item $F_{\alpha} \supset F_{\beta}$, and \hspace*{\fill} (specificity) 
                \item $\not \exists \dynamiccasegamma  \in D$ with \hspace*{\fill} (concision)
                \begin{enumerate}[i]
		          \item either  $F_{\alpha} \supset F_{\gamma} \supset F_{\beta}$,
		          \item or $F_{\gamma} = F_{\alpha}$ and $S_{\alpha} \sqsupset S_{\gamma}$,
		          \item or $F_{\beta} = F_{\gamma}$ and $S_{\alpha} \sqsupseteq S_{\gamma} \sqsupset S_{\beta}$;
                \end{enumerate}
            \end{enumerate}
            \item or
                \begin{enumerate}
                    \item $F_{\beta} = F_{\alpha}$ and $S_{\alpha} \sqsupset S_{\beta}$ and \hspace*{\fill} (advance)
                    \item $\not \exists(F_\gamma, S_{\gamma}, y_{\alpha}) \in D$ with $F_{\gamma} = F_{\alpha}$\\and $S_{\alpha} \sqsupset S_{\gamma} \sqsupset S_{\beta}$; \hspace*{\fill} (proximity)
                \end{enumerate}
        \end{enumerate}

    \item for $\dynamiccasealpha \!\in \!D \!\cup \!\{\dynamiccasedefault\}$, it holds that $\dynamiccasenew \!\rightsquigarrow\! \dynamiccasealpha$ iff either $F_{N} \!\not \supseteq \!F_{\alpha}$ or $S_{N} \!\not \sqsupseteq\! S_{\alpha}$.
    
\end{itemize}

\end{definition}


\adamnew{Intuitively, a case $\dynamiccasealpha$ attacks a case $\dynamiccasebeta$ iff their outcomes differ (condition \textbf{1}) and one of the following conditions is met. Firstly, $\dynamiccasealpha$ is more specific than $\dynamiccasebeta$ and is the most concise such case. If there are multiple such cases, $\dynamiccasealpha$ must be the closest by stages (condition~\textbf{2}). Secondly, if $\dynamiccasealpha$ has advanced further and changed the outcome, it suggests that $\dynamiccasebeta$ might be missing some features expected to be acquired as it progresses through the stages. Proximity identifies the earliest stage when the outcome change is expected (condition \textbf{3}).}


\section{Motivations}
\label{section:motivation}

The partial order used to define attacks in the casebase is a key choice affecting the performance and explanations generated by an AA-CBR model. \adamnew{Selecting a different partial order at model construction may result in varying model performance even when using the same dataset. Additionally, different partial orders have different semantic meanings, so the explanations will differ when describing how an AA-CBR model made its final prediction.} However, using a single partial order limits users' ability to express domain-specific preferences over how cases are compared.

\begin{figure}[t!]
    \begin{subfigure}[t]{0.48\textwidth}
        \centering
        \resizebox{0.55\textwidth}{!}{
        \begin{tikzpicture}[main/.style = {draw, rectangle}]
    
            \node[main, label=left:$C_0$] (C0) at (0, 0)  {$(\emptyset, - )$};
            \node[main, label=left:$C_1$] (C1) at (-3, 2) {$(\{a, b\}, + )$};
            \node[main, label=left:$C_2$] (C2) at (0, 2)  {$(\{c\}, + )$};
            \node[main, label=left:$C_3$] (C3) at (-3, 4)  {$(\{d\}, - )$};
            \node[main, label=above:$N_1$] (N) at (0, 4)  {$(\{a, b, d\}, ?)$};
    
            \draw[-{Latex[length=2mm, width=3mm]}] (C1) -- (C0);                   
            \draw[-{Latex[length=2mm, width=3mm]}] (C2) -- (C0);                   
            
            \node[] (l1) at (-4, 0)  {}; 
            \node[] (l2) at (-2, 0) {Attacks}; 
            \draw[-{Latex[length=2mm, width=3mm]}]  (l2) -- (l1);
            
            \node[] (l3) at (-4, 0.5)  {}; 
            \node[] (l4) at (-2, 0.5) {Irrelevance}; 
            
            \begin{scope}[transparency group, opacity=0.5]
                \draw[-{Latex[length=2mm, width=3mm]}, red, dashed] (N) -- (C2);
                \draw[-{Latex[length=2mm, width=3mm]}, red, dashed] (l4) -- (l3);
            \end{scope}
    
        \end{tikzpicture} 
        }
        \caption{With $\succcurlyeq = \supseteq$ and $\nsim = \not \supseteq$, the prediction for $N_{1}$ is $+$, ignoring case $C_{3}$, which a clinician might expect to be used if  feature $d$ is considered more important for this task.}
        \label{fig:aa-cbr-motivating-example-1}
    \end{subfigure}
    \hfill
    \begin{subfigure}[t]{0.48\textwidth}
        \centering
        \resizebox{0.55\textwidth}{!}{
        \begin{tikzpicture}[main/.style = {draw, rectangle}]
    
            \node[main, label=left:$C_0$] (C0) at (0, 0)  {$(\emptyset, \emptyset, - )$};
            \node[main, label=left:$C_1$] (C1) at (-3, 2) {$(\emptyset, \{a, b\}, + )$};
            \node[main, label=left:$C_2$] (C2) at (0, 2)  {$(\{c\}, \emptyset, + )$};
            \node[main, label=left:$C_3$] (C3) at (-3, 4)  {$(\{d\}, \emptyset, - )$};
            \node[main, label=above:$N_1$] (N) at (0, 4)  {$(\{d\}, \{a, b\}, ?)$};

            \draw[-{Latex[length=2mm, width=3mm]}] (C1) -- (C0);                   
            \draw[-{Latex[length=2mm, width=3mm]}] (C3) -- (C1);

            \begin{scope}[transparency group, opacity=0.5]
                \draw[-{Latex[length=2mm, width=3mm]}, red, dashed] (N) -- (C2);
            \end{scope}
    
        \end{tikzpicture} 
        }
        \caption{With $\succcurlyeq = \succcurlyeq_{lex}$ and $\nsim = \nsim_{lex}$, the prediction for $N_{1}$ is $-$, where $C_{3}$ is now instrumental in this classification.}
        \label{fig:aa-cbr-motivating-example-2}
    \end{subfigure}
    \begin{subfigure}[t]{0.48\textwidth}
        \centering
        \resizebox{0.55\textwidth}{!}{
        \begin{tikzpicture}[main/.style = {draw, rectangle}]
    
            \node[main, label=left:$C_0$] (C0) at (0, 0)  {$(\emptyset, \emptyset, - )$};
            \node[main, label=left:$C_1$] (C1) at (-3, 2) {$(\emptyset, \{a, b\}, + )$};
            \node[main, label=left:$C_2$] (C2) at (0, 2)  {$(\{c\}, \emptyset, + )$};
            \node[main, label=left:$C_3$] (C3) at (-3, 4)  {$(\{d\}, \emptyset, - )$};
            \node[main, label=above:$N_2$] (N) at (0, 4)  {$(\{c\}, \{a\}, ?)$};

            \draw[-{Latex[length=2mm, width=3mm]}] (C1) -- (C0);                   
            \draw[-{Latex[length=2mm, width=3mm]}] (C3) -- (C1);

            \begin{scope}[transparency group, opacity=0.5]
                \draw[-{Latex[length=2mm, width=3mm]}, red, dashed] (N) -- (C3);
                \draw[-{Latex[length=2mm, width=3mm]}, red, dashed] (N) -- (C1);
            \end{scope}
    
        \end{tikzpicture} 
        }
        \caption{With $\succcurlyeq = \succcurlyeq_{lex}$ and $\nsim = \nsim_{lex}$, the prediction for $N_{2}$ is $-$, despite $c$ being a high-priority feature associated with outcome $+$ used in case $C_{2}$, which is ignored in this classification.}
        \label{fig:aa-cbr-motivating-example-4}
    \end{subfigure}

\caption{A comparison of scenarios utilising AA-CBR with a naive attempt to introduce preferences}
\label{}
\end{figure}
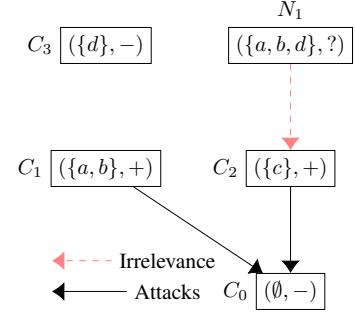
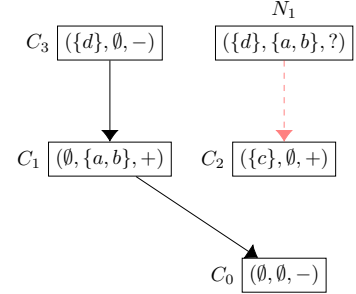
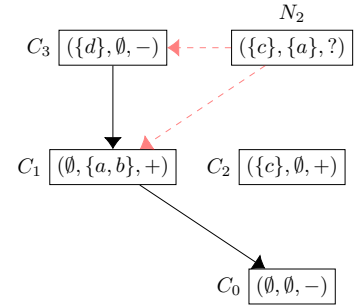

\begin{example}
\label{example:aa-cbr-motivating-1}
    Consider a simple medical assessment tool for assessing patients' well-being. The tool represents patient features using sets and classifies patients as either being in `good' health $(+)$ or `poor' health $(-)$. 
    
    A past patient $(C_1)$ who gets regular exercise (feature $a$) and has a healthy diet (feature $b$) is in good health is represented by the case $(\{a, b\}, +)$. Similarly, a patient $(C_2)$ who has adequate sleep (feature $c$) is represented by the case $(\{c\}, +)$. In contrast, a patient $(C_3)$ who currently has an infection (feature $d$) and is considered in poor health is represented by the case $(\{d\}, -)$. If we have no information from a patient, we act cautiously and assume they have a poor health outcome, so the default case $(C_0)$ is $(\emptyset, -)$.     

    Figure~\ref{fig:aa-cbr-motivating-example-1} shows the AF derived from AA-CBR to classify new case $N_1$ with features $ \{a, b, d\}$. Here, we use $\succcurlyeq = \supseteq$ and $\nsim = \not \supseteq$ as in~\cite{aa-cbr} \adamnew{where casebase cases only attack those with an opposing outcome and a subset of features.} \adamnew{$N_{1}$ attacks $C_{2}$ by irrelevance as $C_{2}$ presents with feature $c$, which $N_{1}$ does not present with.} Case $C_{1}$ is unattacked, \adamnew{and attacks $C_{0}$}, so the default case is not accepted. \adamnew{The grounded extension is $\{ N_{1}, C_{3}, C_{1} \}$}. As a result, we classify the new case with outcome~$+$. Case $C_{3}$ is not used in this classification, and thus neither is feature $d$.    
\end{example}

The outcome in Example~\ref{example:aa-cbr-motivating-1} may be counter-intuitive to a clinician deeming that having an infection $(d)$ and sleeping well $(c)$ are more important for this health assessment. Thus, despite regular exercise $(a)$ and a healthy diet $(b)$, the clinician may wish for the new case to be classified as having a poor health outcome.  Because $C_{3}$ contains an important feature ($d$), the clinician  
would expect the classification to utilise this case. Introducing preferences over constituent parts of the argument structure can thus change the ordering of the arguments to achieve the desired behaviour. The partial order used can be modified to create preferences between groups of features.


    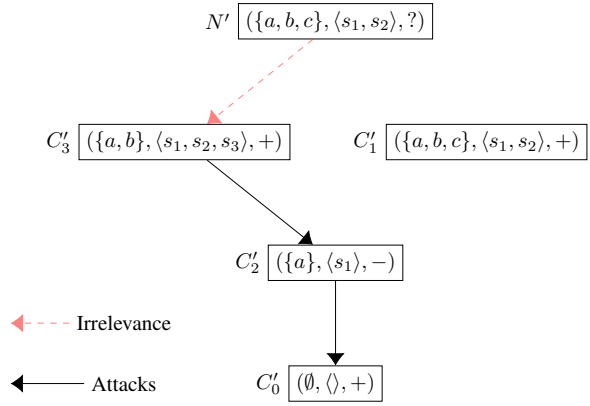
\begin{figure}[t!]
        \centering
    \resizebox{.45\textwidth}{!}{
    \begin{tikzpicture}[node distance={0}, decoration={bent,aspect=.3}, main/.style = {draw, rectangle}]

        \node[main, label=left:$C'_0$] (C0) at (2.5, 0)  {$(\emptyset, \langle \rangle, + )$}; 
        \node[main, label=left:$C'_1$] (C1) at (5, 4)  {$(\{a, b, c\}, \langle s_1, s_2 \rangle, +)$}; 
        \node[main, label=left:$C'_2$] (C2) at (2.5, 2)  {$(\{a\}, \langle s_1 \rangle, -)$}; 
        \node[main, label=left:$C'_3$] (C3) at (0, 4) {$(\{a, b\}, \langle s_1, s_2, s_3 \rangle, +)$}; 
        \node[main, label=left:$N'$] (N) at (2.5, 6) {$(\{a, b, c\}, \langle s_1, s_2 \rangle, ?)$};

        \draw[-{Latex[length=2mm, width=3mm]}] (C3) -- (C2);
        \draw[-{Latex[length=2mm, width=3mm]}] (C2) -- (C0);

        \node[] (l1) at (-3, 0)  {}; 
        \node[] (l2) at (-1, 0) {Attacks}; 
        \draw[-{Latex[length=2mm, width=3mm]}]  (l2) -- (l1);
        
        \node[] (l3) at (-3, 1)  {}; 
        \node[] (l4) at (-1, 1) {Irrelevance};

        \begin{scope}[transparency group, opacity=0.5]
            \draw[-{Latex[length=2mm, width=3mm]}, red, dashed] (N) -- (C3);
            \draw[-{Latex[length=2mm, width=3mm]}, red, dashed] (l4) -- (l3);
        \end{scope}

    \end{tikzpicture} 
    }
    \caption{An AF derived from AA-CBR with Stages as defined in Definition~\ref{def:aa-cbr-with-stages}. Despite $C'_1$ having features and stages equal to $N'$, the outcome predicted for $N'$ is $-$, which differs from $C'_1$'s outcome. The nearest cases property does not hold.}
    \label{fig:aa-cbr-nearest-fails}
    \end{figure}

\begin{example}
    \label{example:aa-cbr-motivating-1b}
    Let us split the characterisation of each data point into two sets: a high-priority feature set and a low-priority feature set.\footnote{This split is for presentation purposes only - each example contains the same features as in Example~\ref{example:aa-cbr-motivating-1}.} Between two cases, the one that has a superset of high-priority features attacks, but if the two cases have the same high-priority feature set, then the one that has a superset of low-priority features attacks. This lexicographic application can be achieved by a single partial order defined as follows:
    
    Let $X \subseteq \mathbb{F}_{H} \times \mathbb{F}_{L}$ where $\mathbb{F}_{H} = \{ c, d \}$ is the set of possible high-priority features and $\mathbb{F}_L = \{a, b\}$ is the possible set of low-priority features. A case is of the form $((H, L), y)$ where $H$ contains the high-priority features and $L$ the low-priority features. Throughout this paper, we will remove the inner brackets when presenting cases for clarity. 
    
    Let $\hlxalpha \supseteq_{H} \hlxbeta$ iff $H_{\alpha} \supseteq H_{\beta}$ and $\hlxalpha \supseteq_{L} \hlxbeta$ iff $L_{\alpha} \supseteq L_{\beta}$. The AA-CBR model can then be instantiated with a lexicographic application of these orders, $\succcurlyeq_{lex}$, where $\hlxalpha \succcurlyeq_{lex} \hlxbeta \text{ iff }$
    \begin{enumerate}
        \item $\hlxalpha \supset_{H} \hlxbeta \text{ or }$
        \item $\hlxalpha =_{H} \hlxbeta \text{ and } \hlxalpha \supseteq_{L} \hlxbeta$.
    \end{enumerate}

    We define the notion of irrelevance such that the attacks by the new case are the same as in Example~\ref{example:aa-cbr-motivating-1}: 
    
    \noindent \mbox{$\hlxalpha \nsim_{lex} \hlxbeta$ iff $H_{\alpha} \not \supseteq H_{\beta}$ or $L_{\alpha} \not \supseteq L_{\beta}$} 

    The AA-CBR model is instantiated with $\succcurlyeq = \succcurlyeq_{lex}$ and $\nsim = \nsim_{lex}$ and the cases in Example~\ref{example:aa-cbr-motivating-1} are modified moving features to their respective priority feature sets. 
    
    The resulting AF is shown in Figure~\ref{fig:aa-cbr-motivating-example-2}. Now $C_{3}$ defends the default case because $d$ is a higher priority feature not present in $C_{1}$. \adamnew{The grounded extension is thus $\{ N_{1}, C_{3}, C_{0}\}$.} The classification for $N_{1}$ is therefore $-$, as desired. 

    However, $C_{1}$ is now considered a more concise case (in the sense of Definition~\ref{def:aa-cbr_geq}, item~\ref{def:concision}) to the default case compared to $C_{2}$, preventing $C_{2}$ from attacking the default case. If we had another new case, $N_{2}$, with features $\{a, c\}$, the expected outcome should be $+$, as both these features are only ever associated with this outcome in the casebase. Furthermore, $C_{2}$ should contribute to the classification as $c$ is a high-priority feature. This situation is modelled in Figure~\ref{fig:aa-cbr-motivating-example-4}. With $\succcurlyeq_{lex}$, $C_2$ is not used, and again, a high-priority feature is ignored. \adamnew{The grounded extension is $\{ N_{2}, C_{2}, C_{0} \}$}, resulting in the counter-intuitive outcome,~$-$, predicted for~$N_{2}$.
\end{example}

This illustrates how AA-CBR cannot effectively consider preferences over constituent parts of an argument. We require a new form of AA-CBR that respects these preferences by definition. Such a model should allow for an arbitrary number of preferences to be defined with no restrictions on how cases are characterised.

Definition~\ref{def:aa-cbr-with-stages} showcases a notion of preferences over comparisons of cases which introduced stages. This approach, however, is not generalised to work with characterisations beyond features and stages, nor allows for an arbitrary number of preferences to be defined. Furthermore, a useful property of AA-CBR \adamnew{relies on} \textit{nearest cases}, \adamnew{which are cases that are ``closest'' by $\succcurlyeq$}\footnote{We will provide a formal definition of nearest cases in our setting in Section~\ref{section:properties}.}: if all cases that are nearest to the new case agree on an outcome, then this is the outcome predicted. Whilst no definition of nearest cases is provided for AA-CBR with Stages in~\cite{arbitrated-argumentative-dispute}, we highlight an example where this property would clearly be violated. 

\begin{example}
\label{example:nearest-neighbour}






Consider the casebase $D' = \{C'_{0}, C'_{1}, C'_{2}, C'_{3}\}$. The AF derived from AA-CBR with Stages for new case $N'$ is shown in Figure~\ref{fig:aa-cbr-nearest-fails}. Note that the features and stages of $C'_{1}$ are the same as in $N'$. We, therefore, expect the predicted outcome for $N'$ to be the same ($+$). However, this is not the case. \adamnew{The grounded extension is $\{ N', C'_{1}, C'_{2} \}$} and the predicted outcome for $N'$~is~$-$. Clearly, the nearest case condition does not hold.
\end{example}

Consequently, we cannot simply generalise the approach in AA-CBR with Stages. We need a new method that meets the requirements for allowing preferences to be defined and obeyed while ensuring that the nearest case property holds.

    
    
    
            
            
    
    

\section{Preference-Based Abstract Argumentation for Case-Based Reasoning}
\label{section:methodology}


We introduce \aacbrp~and \adamnew{its} regular variant \raacbrp, where a user can employ various preorders\footnote{
Previous incarnations of AA-CBR use a partial order; as we use multiple orders to compare cases, each can be a preorder. For example, we have $(\{c,d\}, \{a\}) \supseteq_{H} (\{c,d\}, \{b\})$ and $(\{c,d\}, \{b\}) \supseteq_{H} (\{c,d\}, \{a\})$ but these two characterisations are not equivalent.
}
defined over constituent parts of an argument. The preorders are sorted by preferences and applied using a lexicographic strategy. \adamnew{We begin by defining the collection of preorders.}


\begin{definition}[Preference Ordering]
    Let \mbox{$\pseq$ = $\langle \succcurlyeq_{1}, \ldots, \succcurlyeq_n \rangle$} be a sequence of preorders, each defined over $X$. Each preorder, $\succcurlyeq_{i}$, is a reflexive and transitive relation, with a corresponding strict preorder, $\succ_{i}$ (which is irreflexive and transitive), and an equivalence relation, $=_{i}$ (which is reflexive, symmetric, and transitive). For cases $\casealpha, \casebeta  \in D \cup \{\casedefault \}$, we define:
    \begin{itemize}

        \item $\xalpha \succ_{i} \xbeta $ iff $\xalpha \succcurlyeq_{i} \xbeta$ and $\xbeta \not \succcurlyeq_{i} \xalpha$  

        \item $\xalpha =_{i} \xbeta$ iff $\xalpha \succcurlyeq_{i} \xbeta$ and $\xbeta \succcurlyeq_{i} \xalpha$

    \end{itemize}

\end{definition}

\begin{figure*}[t]
    \hfill
    \begin{subfigure}[t]{.45\textwidth}
        \centering
        \resizebox{0.7\textwidth}{!}{
        \begin{tikzpicture}[> = stealth,  shorten > = 1pt,   auto,   node distance = 1.3cm, main/.style={draw, rectangle}]
            \node[] (d) [          ] {};
            \node[main, label=above:$\alpha$] (a) [left  of=d] {\tiny$(\{c, d\}, \{a, b\}, +)$};
            \node[main, label=above:$\gamma$] (g) [right of=d] {\tiny$(\{c, d\}, \{a\}, +)$};
            \node[main, label=below:$\beta$] (b) [below of=d] {\tiny$(\emptyset, \emptyset, -)$};

            \path[->](a) edge node (m) {\scriptsize$\supseteq_{H}$}  (b);
            \path[->,     ] (g) edge node  {\scriptsize$\supseteq_{H}$} (b);
            \draw[red,shift={(m)}](-0.4,-0.4)--(0.1,-0.25);
        \end{tikzpicture}
        }
        \caption{$\gamma$ is equivalent to $\alpha$ by order $\supseteq_{H}$ and smaller by order $\supseteq_{L}$, blocking the attack from $\alpha$.}
        \label{fig:blocked-attack-a}
    \end{subfigure}
    \hfill
    \begin{subfigure}[t]{.45\textwidth}
        \centering
        
        \resizebox{0.7\textwidth}{!}{
        \begin{tikzpicture}[> = stealth,  shorten > = 1pt,   auto,   node distance = 1.3cm, main/.style={draw, rectangle}]
            \node[] (d) [          ] {};
            \node[main, label=above:$\alpha$] (a) [left  of=d] {\tiny$(\{c, d\}, \{a, b\}, +)$};
            \node[main, label=above:$\gamma$] (g) [right of=d] {\tiny$(\emptyset, \{a, b\}, +)$};
            \node[main, label=below:$\beta$] (b) [below of=d] {\tiny$(\emptyset, \emptyset, -)$};

            \path[->](a) edge node (m) {\scriptsize$\supseteq_{H}$}  (b);
            \path[->,     ] (g) edge node  {\scriptsize$\supseteq_{L}$} (b);
            \draw[red,shift={(m)}](-0.4,-0.4)--(0.1,-0.25);
        \end{tikzpicture}
        }
        \caption{$\gamma$ is smaller than $\alpha$ and equivalent to $\beta$ by order $\supseteq_{H}$ but greater than $\beta$ on order $\supseteq_{L}$, thus blocking the attack from $\alpha$.}
        \label{fig:blocked-attack-b}
    \end{subfigure}
    \caption{Two AFs showcasing potential attacks that are blocked by more concise cases according to the two concision conditions}
    \label{fig:blocked-attack}
\end{figure*}
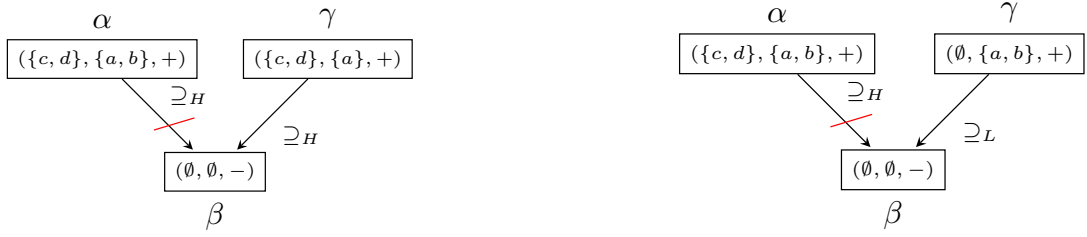

\FT{In simple terms, the} ordering of $\pseq$ determines the preferences of the preorders. If two cases are equivalent by the first preorder, then a comparison under the second order can be used. If they are also equivalent by the second order, the third can be used. This process continues, using subsequent preorders when equivalence occurs. By incorporating multiple methods for comparing cases, we afford flexibility to select the approach that optimises performance and injects domain relevance. Consequently, we can create a new AA-CBR-based model that utilises $\pseq$. To do so, we make use of the following shorthand:

\begin{notation}[]
We use $\xalpha \bigsucccurlyeq{j}{k} \xbeta$, where $j \leq k$, to mean $\xalpha$ is equivalent or larger than $\xbeta$ on all preorders between orders $\succcurlyeq_{j}$ and $\succcurlyeq_{k}$ (inclusive). 
Formally:
\begin{align*}
\xalpha \bigsucccurlyeq{j}{k} \xbeta \text{ iff } \forall i \in [j, k], \xalpha \succcurlyeq_{i} \xbeta. 
\end{align*}

\noindent
Shorthands for $\bigeq{j}{k}$ and $\bigsucc{j}{k}$ are defined analogously.

\end{notation}

\adamnew{With this, we can create a definition similar to Definition \ref{def:aa-cbr_geq}} that enforces attacks in the direction defined by the \adamnew{sequence of} partial orders and that attacks only originate from the most concise case possible, thus representing minimal change between cases. With the addition of multiple preorders, the concision condition must now consider the lexicographic application of the orders. We first define a notion of \emph{potential attacks} for a particular order $\succcurlyeq_{i}$, which describes attacks that would transpire if we do not consider concision.

\begin{definition}[Potential Attacks] Let $\alpha = \casealpha$ and $\beta = \casebeta$. For $\alpha, \beta \in D \cup \{ \casedefault \}$, we define a \emph{potential attack} on order $\succcurlyeq_{i}$ as: $\alpha \potentialattacks{i} \beta$ iff

    \begin{enumerate}[i]
        \item $y_{\alpha} \neq y_{\beta}$, and
        \item $\xalpha \succ_i \xbeta$, and
        \item $\xalpha \bigeq{1}{i-1} \xbeta$.
    \end{enumerate}
\label{def:potential-attacks}
\end{definition}

\FT{Intuitively,} $\alpha$ potentially attacks $\beta$ on order $\succcurlyeq_{i}$ when the outcomes of the cases are different (\textbf{i}), $\xalpha$ is strictly greater than $\xbeta$ by order $\succ_{i}$ (\textbf{ii}) and they are equivalent on all orders before $\succ_{i}$ (\textbf{iii}). Conditions (\textbf{ii}) and (\textbf{iii}) apply $\pseq$ lexicographically. We can subsequently restrict attacks to the most concise attacks.

\begin{definition}[Casebase Attacks] 
Let $\alpha = \casealpha$ and $\beta = \casebeta$.
For $\alpha, \beta \in D \cup \{ \casedefault \}$, we define an attack on order $\succcurlyeq_{i}$: $\alpha \attacks[i] \beta$ iff
    \label{def:specificity-attacks}

    \begin{enumerate}[i]
        \item $\alpha \potentialattacks{i} \beta$ and
        \item $\nexists \gamma = \casegamma \in D \cup \{ \casedefault \}$ with $\xalpha \bigsucccurlyeq{1}{n} \xgamma$ and
              \begin{enumerate}
              
                  \item either $\gamma \potentialattacks{i} \beta$ and $\exists l \geq i,$ $\xalpha \succ_{l} \xgamma$, 
                  
                  \item or $\gamma \not \potentialattacks{i} \beta$ and $\exists l > i,$ $\gamma \potentialattacks{l} \beta$. 
                  
              \end{enumerate}
    \end{enumerate}
\end{definition}

\adamnew{This} definition states that case $\alpha$ attacks case $\beta$ on order~$\succcurlyeq_{i}$ if two conditions are met. Firstly, $\alpha$ potentially attacks $\beta$ on order $\succcurlyeq_{i}$ (\textbf{i}), and secondly, $\alpha$ is the most concise case capable of such an attack (\textbf{ii}). We illustrate the two conditions of concision in Figure~\ref{fig:blocked-attack}, utilising the same characterisation approach as Example~\ref{example:aa-cbr-motivating-1b} and letting $\pseq = \langle \supseteq_{H}, \supseteq_{L} \rangle$.

\begin{enumerate}

\item[(a)] Condition \textbf{a} details the circumstance where both $\alpha$ and $\gamma$ can potentially attack $\beta$ on order $\succcurlyeq_{i}$ and $\xgamma$ is smaller than $\xalpha$ on at least one of the orders in $\pseq$. As both $\alpha$ and $\gamma$ are potential attackers of $\beta$, we know that they are both equivalent to $\beta$ on all orders up to and including~$\succcurlyeq_{i-1}$, that is $\xalpha \bigeq{1}{i-1} \xgamma \bigeq{1}{i-1} \xbeta$. Furthermore, as $\xalpha \bigsucccurlyeq{1}{n} \xgamma$, we know there is an $l'$ such that $l \geq l' \geq i$ in which $\xalpha \bigeq{i - 1}{l' - 1} \xgamma$ and $\xalpha \succ_{l'} \xgamma$. This is to say, both $\xalpha$ and $\xgamma$ are equivalent on all orders up to some order $\succcurlyeq_{l'}$ in which $\xalpha$ is greater than $\xgamma$. As a result, we can take $\gamma$ as the more concise case. In Figure~\ref{fig:blocked-attack-a}, $\alpha$ potentially attacks $\beta$ by $\supseteq_{H}$ but is blocked by the more concise case $\gamma$ because they are equivalent on the first preorder,~$\supseteq_{H}$,  i.e. both have features $\{c, d\}$, but $\gamma$ is less specific on the second preorder as $\{a, b\} \supseteq \{a\}$.


\item[(b)] Condition \textbf{b} covers the instance where $\alpha$ is greater than $\gamma$ on $\succcurlyeq_{i}$ but $\gamma$ cannot potentially attack $\beta$ on order $\succcurlyeq_{i}$ as $\gamma$ and $\beta$ are equivalent on all orders up to and including~$\succcurlyeq_{i}$. $\gamma$ is, therefore, only a more concise case if it can potentially attack by some order $\succcurlyeq_{l}$ where $l>i$. This is because, by $\xalpha \bigsucccurlyeq{1}{n} \xgamma$, we have that $\alpha$ is greater than $\gamma$ on all orders up to an including order $\succcurlyeq_{l}$ that is $\xalpha \bigsucccurlyeq{1}{l} \xgamma$. Therefore, we know that $\xalpha \succcurlyeq_{l} \xgamma \succ_{l} \xbeta$ and thus can take $\gamma$ as a more concise case. In Figure~\ref{fig:blocked-attack-b} $\alpha$ potentially attacks $\beta$ by $\supseteq_{H}$ but is blocked by $\gamma$ which has an equivalent high-priority feature set as $\beta$ (both of which are $\emptyset$) and $\gamma$ has a superset of $\beta$'s low-priority features so $\gamma$ attacks by order~$\supseteq_{L}$.

\end{enumerate}



\adamnew{Furthermore,} the constraint $\xalpha \bigsucccurlyeq{1}{n} \xgamma$ in (\textbf{ii}, \adamnew{Definition~\ref{def:specificity-attacks}}) ensures that the attacker $\alpha$ can never be blocked by a case that is greater by any preorder. If this did not hold, a nearest case may be irrelevant to a new case as in Example~\ref{example:nearest-neighbour}. This is discussed further in Section~\ref{section:properties}. 

However, \adamnew{Definition \ref{def:specificity-attacks}} does not cover the circumstance where two cases may be equivalent to each other on all orders in $\pseq$, leading to an \emph{incoherent} casebase. 

\begin{definition}[Coherent and Incoherent Casebases]

A casebase $D$ is \emph{coherent} iff there are no two cases $\casealpha, \casebeta \in D$ such that $\xalpha \bigeq{1}{n} \xbeta$ and $y_{\alpha} \not = y_{\beta}$, and it is \emph{incoherent} otherwise.
\end{definition}

Incoherence is handled with symmetric attacks:

\begin{definition}[Incoherent Attacks] 
\label{incoherent-attacks}
For $\casealpha, \casebeta \in D \cup \{\casedefault\}$, we define an incoherent attack: $\casealpha \incoherentattacks \casebeta$ iff $\xalpha \bigeq{1}{n} \xbeta$ and $y_{\alpha} \neq y_{\beta}$.

\end{definition}

We have now defined all possible attacks between arguments in the casebase. We add the new case to the AF analogous to Definition~\ref{def:aa-cbr_geq}, utilising a notion of irrelevance.

\begin{definition}[New Case Attacks] 
\label{new-case-attacks}
For $\casealpha \in D \cup \{\casedefault\}$ and new case $N = \casenew$, we define a new case attack: $N \newattacks \casealpha$ iff $N \nsim \casealpha.$

\end{definition}

With these attacks defined, we can now define the argumentation framework for \aacbrp.

\begin{definition}[\FT{Argumentation Framework drawn from $\pseq$}] The argumentation framework \afp{N} mined from dataset $D$ and a new case $N$, with a sequence of preorders $\pseq$ of length $n$ is $(Args, \attacks)$ where
\label{def:aa-cbr-p}

\begin{itemize}
    \item[] $Args = D \cup \{\casedefault\} \cup \{N\}$ and
    \item[] \mbox{$\attacks = (\bigcup_{i=1}^{n} \attacks[i]) \cup \incoherentattacks \cup \newattacks$}.
\end{itemize}

As with Definition~\ref{def:aa-cbr-prediction}, the outcome predicted for $N$, written as \faacbrp{N}, is  $\delta$ if $\casedefault$ is in the grounded extension $\mathbb{G}$ of the AF and $\bar{\delta}$ otherwise. 
\end{definition}

We \FT{also} define a \emph{regular} variant of \aacbrp~that enforces that irrelevance is written in terms of the partial orders in $\pseq$ and that $\x{\delta}$ is the least element by all orders in $\pseq$. 
\adamnew{The regular variant naturally aligns with expectations that irrelevance is directly related to the partial order defining attacks. Moreover, regular \aacbrp~has desired properties that we discuss in Section \ref{section:properties}.}

\begin{definition}[Regular]
\label{def:regular-aa-cbr-p}
    The AF mined from $D$ and $N$, with default argument $\casedefault$ is \emph{regular} when:

    \begin{enumerate}
        \item The irrelevance relation $\nsim$ is defined as $\xalpha \nsim \xbeta$ iff $\exists i, \xalpha \not \succcurlyeq_{i} \xbeta$.
        \item $\x{\delta}$ is the least element on all orders in $\pseq$.
    \end{enumerate}

\end{definition}

\adamnew{Definition \ref{def:regular-aa-cbr-p}} is a generalisation of the approach in~\cite{monotonicity-and-noise-tolerance}. We will use the notation \raacbrp~to refer to the regular variant of \aacbrp~with an arbitrary instantiation of $\pseq$. We will populate the subscript with a sequence of preorders to refer to an instantiation of the model. Similarly, we will use \rafp{N}~to refer to the regular variant of \afp{N}. \adamnew{To illustrate, consider the following:} 

\begin{figure}
    
    \begin{subfigure}[t]{0.5\textwidth}
        \centering
        \resizebox{0.5\textwidth}{!}{
        \begin{tikzpicture}[main/.style = {draw, rectangle}]
    
            \node[main, label=left:$C_0$] (C0) at (0, 0)  {$(\emptyset, \emptyset, - )$};
            \node[main, label=left:$C_1$] (C1) at (-3, 2) {$(\emptyset, \{a, b\}, + )$};
            \node[main, label=left:$C_2$] (C2) at (0, 2)  {$(\{c\}, \emptyset, + )$};
            \node[main, label=left:$C_3$] (C3) at (-3, 4)  {$(\{d\}, \emptyset, - )$};
            \node[main, label=above:$N_1$] (N) at (0, 4)  {$(\{d\}, \{a, b\}, ?)$};

            \draw[-{Latex[length=2mm, width=3mm]}] (C1) -- (C0);                   
            \draw[-{Latex[length=2mm, width=3mm]}] (C3) -- (C1);                   
            \draw[-{Latex[length=2mm, width=3mm]}] (C2) -- (C0);

            \begin{scope}[transparency group, opacity=0.5]
                \draw[-{Latex[length=2mm, width=3mm]}, red, dashed] (N) -- (C2);
            \end{scope}
    
        \end{tikzpicture} 
        }
    \caption{With $\raacbrp[\supseteq_{H}, \supseteq_{L}]$, the outcome predicted for $N_{1}$ is as expected, following the fact that ${d}$ is a high-priority feature.}
    \label{fig:aa-cbr-methodology-example-1}
    \end{subfigure}

    \begin{subfigure}[t]{0.5\textwidth}
        \centering
        \resizebox{0.5\textwidth}{!}{
        \begin{tikzpicture}[main/.style = {draw, rectangle}]
    
            \node[main, label=left:$C_0$] (C0) at (0, 0)  {$(\emptyset, \emptyset, - )$};
            \node[main, label=left:$C_1$] (C1) at (-3, 2) {$(\emptyset, \{a, b\}, + )$};
            \node[main, label=left:$C_2$] (C2) at (0, 2)  {$(\{c\}, \emptyset, + )$};
            \node[main, label=left:$C_3$] (C3) at (-3, 4)  {$(\{d\}, \emptyset, - )$};
            \node[main, label=above:$N_2$] (N) at (0, 4)  {$(\{c\}, \{a\}, ?)$};

            \draw[-{Latex[length=2mm, width=3mm]}] (C1) -- (C0);                   
            \draw[-{Latex[length=2mm, width=3mm]}] (C3) -- (C1);                   
            \draw[-{Latex[length=2mm, width=3mm]}] (C2) -- (C0);

            \begin{scope}[transparency group, opacity=0.5]
                \draw[-{Latex[length=2mm, width=3mm]}, red, dashed] (N) -- (C3);
                \draw[-{Latex[length=2mm, width=3mm]}, red, dashed] (N) -- (C1);
            \end{scope}
    
        \end{tikzpicture} 
        }
    \caption{With $\raacbrp[\supseteq_{H}, \supseteq_{L}]$, the outcome predicted for $N_{2}$ is as expected, following the fact that ${c}$ is a high-priority feature.}
        \label{fig:aa-cbr-methodology-example-2}
    \end{subfigure}

    \caption{A comparison of scenarios utilising \aacbrp~to introduce preferences as defined in Example \ref{example:aa-cbr-methodology-1}.}
    
\end{figure}
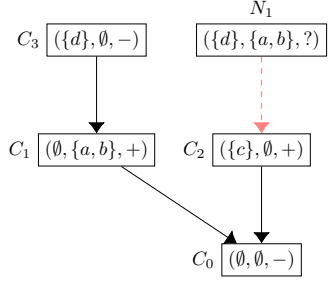
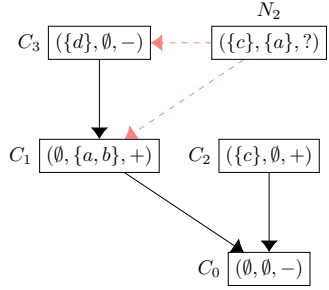

\begin{example} 
\label{example:aa-cbr-methodology-1}
Based on Example~\ref{example:aa-cbr-motivating-1b}, we can utilise \raacbrp~to create a model that obeys preferences in all circumstances. We let \mbox{$\pseq = \langle \supseteq_{H}, \supseteq_{L} \rangle$} and refer to this model as \raacbrp[\supseteq_{H}, \supseteq_{L}]. \rafp[\supseteq_{H}, \supseteq_{L}]{N_{1}} is shown in Figure~\ref{fig:aa-cbr-methodology-example-1} and \rafp[\supseteq_{H}, \supseteq_{L}]{N_{2}} in Figure~\ref{fig:aa-cbr-methodology-example-2}. In both instances, the preferences are respected, and we see that the classification for $N_{1}$ and $N_{2}$ are determined by the most important features, $d$ and $c$, respectively. $C_{1}$ is no longer considered a more concise case than $C_{2}$, thus, they can now both attack~$C_{0}$. \adamnew{The grounded extension for \rafp[\supseteq_{H}, \supseteq_{L}]{N_{1}} is $\{ N_{1}, C_{3}, C_{0} \}$ with the predicted outcome, $-$, as desired. For \rafp[\supseteq_{H}, \supseteq_{L}]{N_{2}} the grounded extension is $\{N_{2}, C_{2}\}$ with the predicted outcome,~$+$, again as desired.}


\end{example}

Example~\ref{example:aa-cbr-methodology-1} illustrates how, in general, AA-CBR by Definition \ref{def:aa-cbr_geq} cannot capture \aacbrp. $C_{2}$ and $C_{3}$ are cases that should be treated with similar importance within the casebase; both have a single high-priority feature but differing outcomes. With a single partial order, as in AA-CBR, we can only support one of two possible scenarios:
\begin{enumerate}
    \item either $C_{2}$ and $C_{3}$ are considered incomparable to $C_{1}$, allowing both $C_{2}$ and $C_{1}$ to attack $C_{0}$ but $C_{3}$ cannot attack $C_{1}$ as in Figure \ref{fig:aa-cbr-motivating-example-1},
    
    \item or $C_{2}$ and $C_{3}$ are considered larger than $C_{1}$ by the partial order, allowing $C_{3}$ to attack $C_{1}$ but then $C_{1}$ is a more concise case than $C_{2}$, meaning only $C_{1}$ can attack $C_{0}$, as in Figure \ref{fig:aa-cbr-motivating-example-2} and \ref{fig:aa-cbr-motivating-example-4}. 
\end{enumerate}
\aacbrp~allows for $C_{2}$ and $C_{3}$ to be treated with similar importance within the casebase, but have $C_{3}$ still attack $C_{1}$ and $C_{2}$ attack $C_{0}$. We can thus conclude the following theorem:


    




\begin{theorem}
\label{theorem:generalised}

\adamnew{There exists a sequence of partial orders $\pseq$ and casebase $D$, such that there is no partial order, $\succcurlyeq$, in which \af{N} constructed with $\succcurlyeq$ is the same as \afp{N} constructed with $\pseq$ and $\faacbr{N}$ is the same as $\faacbrp{N}$.}

    
\end{theorem}

\adamnew{Theorem \ref{theorem:generalised} shows that $\aacbrp$~is capable of constructing AFs that AA-CBR (as defined by Definition \ref{def:aa-cbr_geq}) is not able to and, as a result, predictions made by $\aacbrp$ differ as well. Thus, $\aacbrp$ is a novel contribution that improves on AA-CBR with the use of preferences.}

\section{Nearest and Preferred Cases}
\label{section:properties}

\begin{figure}[t]
    \centering
    \resizebox{.45\textwidth}{!}{
        \begin{tikzpicture}[node distance={0}, decoration={bent,aspect=.3}, main/.style = {draw, rectangle}]

            \node[main, label=left:$C'_0$] (C0) at (2.5, 0)  {$(\emptyset, \langle \rangle, + )$};
            \node[main, label=left:$C'_1$] (C1) at (5, 4)  {$(\{a, b, c\}, \langle s_1, s_2 \rangle, +)$};
            \node[main, label=left:$C'_2$] (C2) at (2.5, 2)  {$(\{a\}, \langle s_1 \rangle, -)$};
            \node[main, label=left:$C'_3$] (C3) at (0, 4) {$(\{a, b\}, \langle s_1, s_2, s_3 \rangle, +)$};
            \node[main, label=left:$N'$] (N) at (2.5, 6) {$(\{a, b, c\}, \langle s_1, s_2 \rangle, ?)$};

            \draw[-{Latex[length=2mm, width=3mm]}] (C3) -- (C2);
            \draw[-{Latex[length=2mm, width=3mm]}] (C1) -- (C2);
            \draw[-{Latex[length=2mm, width=3mm]}] (C2) -- (C0);

            \node[] (l1) at (-3, 0)  {};
            \node[] (l2) at (-1, 0) {Attacks};
            \draw[-{Latex[length=2mm, width=3mm]}]  (l2) -- (l1);

            \node[] (l3) at (-3, 1)  {};
            \node[] (l4) at (-1, 1) {Irrelevance};

            \begin{scope}[transparency group, opacity=0.5]
                \draw[-{Latex[length=2mm, width=3mm]}, red, dashed] (N) -- (C3);
                \draw[-{Latex[length=2mm, width=3mm]}, red, dashed] (l4) -- (l3);
            \end{scope}

        \end{tikzpicture}
    }
    \caption{\rafp[\supseteq, \sqsupseteq]{N} as defined in Example~\ref{example:properties-example-1}. The nearest cases property holds. $C'_{1}$ is not prevented from attacking as $C'_{3}$ is no longer considered a more concise case.}
    \label{fig:aa-cbr-nearest-succeeds}
\end{figure}
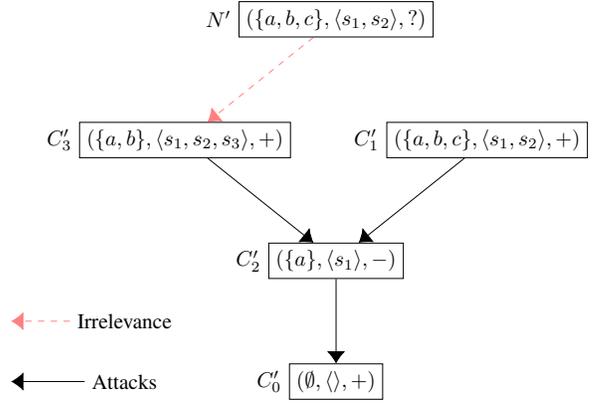

With a coherent casebase, $D$, and a regular AF, we can identify cases which, when in agreement on an outcome, precisely determine the outcome predicted by \aacbrp.
In this section, we will present this formally\footnote{Proofs of 
results can be found in Appendix~\ref{section:proofs}. 
}.
Note that when $D$ is coherent, the argumentation framework is guaranteed to be \emph{acyclic}.

\begin{proposition}[] The argumentation framework corresponding to \raacbrp~is guaranteed to be acyclic for a coherent~$D$.
    \label{prop:acyclic}
\end{proposition}



Previous approaches to AA-CBR have shown that \adamnew{for a coherent casebase} if all cases that are $\textit{nearest}$ to the new case $N$ have the same outcome, then that outcome will be predicted for $N$ \cite{aa-cbr,monotonicity-and-noise-tolerance-appendix}. Here, we generalise this property for \raacbrp.

\begin{definition}[Nearest Case]
    \label{def:nearest-neigbour}

    A case $(\x{\alpha}, y_{\alpha}) \in D \cup \{\casedefault\}$ is \textit{nearest} to $N$ iff $\x{N} \bigsucccurlyeq{1}{n} \x{\alpha}$ and $(\x{\alpha}, y_{\alpha})$ is maximally so, that is, there is no $(\x{\beta}, y_{\beta}) \in D$ such that $\x{N} \bigsucccurlyeq{1}{n} \x{\beta}$ and $\x{\beta} \bigsucccurlyeq{1}{n} \x{\alpha}$ and  $\exists i, \x{\beta} \succ_{i} \x{\alpha}$.

\end{definition}

\adamnew{Intuitively, this} states a case $\casealpha$ that is smaller to $N$ on all orders in $\pseq$ is nearest when there does not exist another case, $\casebeta$, that is similarly smaller to $N$ on all orders, no smaller than $\casealpha$ on any orders and greater than $\casealpha$ on at least one order. If $\casebeta$ exists, it is a better candidate for being the nearest case than $\casealpha$. We can define the relationship between nearest and new cases as follows: 

\begin{theorem}
    \label{theorem:nearest-neighbour}
    If $D$ is coherent and every nearest case to $N$ is of the form $(\x{\alpha}, y)$ for some outcome $y \in Y$, then \rfaacbrp{N} $ = y$.
\end{theorem}

The constraint $\x{\alpha} \bigsucccurlyeq{1}{n} \x{\gamma}$ in (\textbf{ii}) in Definition~\ref{def:specificity-attacks} ensures \adamnew{that Theorem \ref{theorem:nearest-neighbour}} holds. It enforces that a case is only more concise than another if it is smaller by all orders in $\pseq$. Without this constraint, an irrelevant case may prevent a nearest case from contributing to the classification by the concision. This explains the violation of the nearest cases theorem with AA-CBR with Stages in Example~\ref{example:nearest-neighbour}. \raacbrp~resolves this. 

\begin{example}
\label{example:properties-example-1}
To illustrate, we instantiate \raacbrp with $\pseq$ = $\langle \supseteq, \sqsupseteq \rangle$ and case characterisations corresponding to that of Definition~\ref{def:aa-cbr-with-stages}.
For $ \x{\alpha} = (f_{\alpha}, s_{\alpha})$ and \mbox{$\x{\beta} = (f_{\beta}, s_{\beta})$}:  
\begin{itemize}
    \item $\x{\alpha} \supseteq \x{\beta} $ iff $ f_{\alpha} \supseteq f_{\beta}$,  
    \item $\x{\alpha} \sqsupseteq \x{\beta} $ iff $ s_{\alpha} \sqsupseteq s_{\beta}$. 
\end{itemize} 

\noindent We will refer to this model as \raacbrp[\supseteq, \sqsupseteq].

    The argumentation framework, \rafp[\supseteq, \sqsupseteq]{N} corresponding to the cases from Example~\ref{example:nearest-neighbour} is shown in Figure~\ref{fig:aa-cbr-nearest-succeeds}. $C'_{3}$ is not smaller on all orders than $C'_{1}$ and therefore is no longer considered a more concise case. \adamnew{The grounded extension is $\{ N', C'_{1}, C'_{0}\}$} and the predicted outcome is $+$ as expected. As a result, $C'_{1}$ is able to affect the determination of $N'$, which results in the outcome predicted precisely agreeing. The nearest cases property clearly holds.
\end{example}

A modified variant of AA-CBR with Stages as described in Definition~\ref{def:aa-cbr-with-stages} can be derived from \raacbrp[\supseteq, \sqsupseteq] with condition \textbf{2. (b) i} altered to $F_{\alpha} \supset F_{\gamma} \supset F_{\beta}$ and $S_{\alpha} \sqsupseteq S_{\gamma}$. Moreover, by substituting $\pseq = \langle \succcurlyeq \rangle$, we can precisely capture AA-CBR as described in Definition~\ref{def:aa-cbr_geq} by \aacbrp \footnote{Derivations can be found in Appenix~\ref{section:derivations}.}. As a result, our method effectively generalises both of these previous approaches whilst ensuring the nearest case property holds in all circumstances.






However, we can make a stronger claim that preferences themselves are obeyed and used in determining the outcome of a new case.
In Figure~\ref{fig:aa-cbr-methodology-example-1}, both $C_{3}$ and $C_{1}$ are nearest cases but do not agree on an outcome. Yet, the outcome predicted is the same as that for $C_{3}$. This is not incidental; $ C_{3}$ is a \textit{preferred} case.


\begin{definition}[Preferred Case]
    \label{def:preferred-cases}
    A nearest case $\casealpha$ is \textit{preferred} for $N$ if there is no other nearest case $\casebeta$ such that $\exists j,$ $\x{\alpha} \bigeq{1}{j - 1} \x{\beta}$ and $\x{\beta} \succ_{j} \x{\alpha}$.
\end{definition}

Preferred cases are nearest cases that are maximally close to $N$ by the most preferred order possible. As a result, we can prove a similar but stronger condition than that provided by nearest cases: if preferred cases agree on an outcome, then that is the outcome predicted for a new case.

\begin{theorem}
    \label{theorem:preferred-cases}
    If $D$ is coherent and every preferred case to $N$ is of the form $(\x{\alpha}, y)$ for some outcome $y \in Y$ then \rfaacbrp{N} $ = y$.
\end{theorem}

Preferred cases are a subset of nearest cases in which, \adamnew{by Theorem \ref{theorem:preferred-cases}}, the preferred cases alone are sufficient to decide the outcome of a new case when they agree on an outcome.

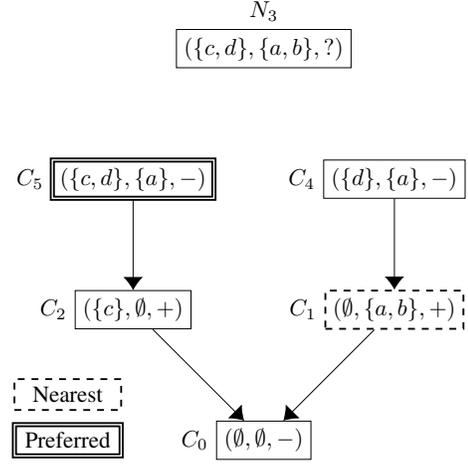
\begin{figure}[t]
    \centering
        \resizebox{0.35\textwidth}{!}{
        \begin{tikzpicture}[main/.style = {draw, rectangle}]
    
            \node[main, label=left:$C_0$] (C0) at (0, 0)  {$(\emptyset, \emptyset, - )$};
            
            \node[main, label=left:$C_2$]                (C2) at (-2, 2) {$(\{c\}, \emptyset, + )$};
            \node[main, thick, dashed, label=left:$C_1$] (C1) at (2, 2)  {$(\emptyset, \{a, b\}, + )$};
            \node[main, thick, double, label=left:$C_5$] (C5) at (-2, 4)  {$(\{c, d\}, \{a\}, - )$};
            \node[main,  label=left:$C_4$]               (C4) at (2, 4)  {$(\{d\}, \{a\}, -)$};
            \node[main, label=above:$N_3$]               (N)  at (0, 6)  {$(\{c, d\}, \{a, b\}, ?)$};
            
            \node[main, thick, double, text width=1.4cm, align=center] (L1) at (-3, 0)  {Preferred};
            \node[main, thick, dashed, text width=1.4cm, align=center] (L2) at (-3, 0.7)  {Nearest};

            \draw[-{Latex[length=2mm, width=3mm]}] (C2) -- (C0);                   
            \draw[-{Latex[length=2mm, width=3mm]}] (C5) -- (C2);                   
            \draw[-{Latex[length=2mm, width=3mm]}] (C1) -- (C0);                   
            \draw[-{Latex[length=2mm, width=3mm]}] (C4) -- (C1);

            \begin{scope}[transparency group, opacity=0.5]
            \end{scope}
    
        \end{tikzpicture} 
        }
    \caption{The argumentation framework as defined in Example \ref{example:preferred-cases}. We see $C_{5}$ is the preferred case and despite disagreeing with nearest case $C_{1}$, the outcome predicted for $N_{3}$ is the same as $C_{5}$, as expected by Theorem~\ref{theorem:preferred-cases}.}
    \label{fig:aa-cbr-p-preferred}
\end{figure}

\begin{example}
\label{example:preferred-cases}

Consider the argumentation framework, \rafp[\supseteq_H, \supseteq_L]{N_3} in Figure \ref{fig:aa-cbr-p-preferred}. $C_{1}$ and $C_{5}$ are each nearest cases to $N_{3}$ as they are maximally close to $N_{3}$ in the casebase as by Definition~\ref{def:nearest-neigbour}. However, these two nearest cases do not agree on an outcome. Only $C_{5}$ is a preferred case; it is maximally close to $N_{3}$ on the high-priority order $\supseteq_{H}$. By Theorem \ref{theorem:preferred-cases}, the outcome of $N_{3}$ is the same as the outcome of $C_{5}$. 

This occurs because $C_{1}$ is attacked by $C_{4}$, which is not considered nearest or preferred but is more concise than $C_{5}$. If $C_{4}$ were missing from this casebase, $C_{5}$ would fill the role and attack $C_{1}$. Therefore, the existence of $C_{5}$ as a preferred case precisely determines the outcome predicted for~$N_{3}$.
    
\end{example}

 Furthermore, by Definitions~\ref{def:potential-attacks} and~\ref{def:specificity-attacks}, \adamnew{we have that if there exists an attack from $\alpha$ to $\beta$, this attack must occur by the most preferred order of $\pseq$ possible.}

\begin{proposition}
    For cases $\alpha, \beta \in D$, if $\exists i, \alpha \attacks[i] \beta$ then $\not \exists k < i, \alpha \attacks[k] \beta$.
\end{proposition}

 This is easy to see since $\alpha \attacks[i] \beta$ implies $\alpha \potentialattacks{i} \beta$, which in turn implies $\alpha \bigeq{1}{i-1} \beta$.

\adamnew{Consequently}, the argumentation framework is constructed by obeying the preference orderings. Moreover, with Theorem \ref{theorem:preferred-cases}, we know that preferred cases can determine the outcome of a new case. We can thus conclude that the outcome predicted for a new case follows the user-defined preferences by construction.




\section{Empirical Evaluation}
\label{section:empirical}
In this section, we showcase how to utilise \aacbrp~models in a medical classification task with preferences defined over features derived from varying data sources. BrainWear~\cite{brainwear-first-42-conference,brainwear-paper} is a study exploring the utility of \emph{physical activity} (PA) data collected via wrist-worn accelerometers in patients with a primary brain tumour. PA data complements traditional questionnaires, that allow patients to report their health status, known as \emph{patient reported outcomes} (PRO). We can predict disease progression in new patients using \aacbrp,~with preferences defined over features derived from PA and PRO data\footnote{
Details on data can be found in Appendix~\ref{section:experiment-methods}.
}.

\subsection{Methodology}


Each completed questionnaire is matched with an 8-week average of PA data centred on the questionnaire date. The objective is to predict if the patient has \emph{progressive disease} or not, as labelled by the outcome of the next MRI scan following the questionnaire or the patient's mortality status. Furthermore, we can incorporate the number of previous instances of progressive disease for each patient, serving as a proxy for cancer progression. We conduct two experiments to evaluate the models. The first experiment solely utilises PA and PRO data, while the second incorporates both PA and PRO data along with the proxy measurement as stages. Each experiment consists of multiple AA-CBR(-$\pseq$) models compared against similarly interpretable baseline models: a decision tree and a K-Nearest Neighbor (kNN).

In the first experiment, we compare AA-CBR with the ${\supseteq}$ relation as defined in Example \ref{example:aa-cbr-motivating-1} (we denote this model as AA-CBR$_\supseteq$)  and AA-CBR with $\succcurlyeq_{lex}$ (denoted AA-CBR$_{\succcurlyeq_{lex}}$) as defined in Example \ref{example:aa-cbr-motivating-1b} where PA and PRO features make up the full suite of features against, \raacbrp[\supseteq_{pa},\supseteq_{pro}] defined analogously to Example~\ref{example:aa-cbr-methodology-1}, where we treat the PA feature set as the high-priority set and the PRO feature set as the low-priority set. 
We have chosen these models to showcase how a naive instantiation of preferences in AA-CBR compares against our more sophisticated approach using \aacbrp. 

For the dynamic models, we compare AA-CBR with Stages as in Definition~\ref{def:aa-cbr-with-stages} using the proxy measurement as the stages. We compare this against \raacbrp[\supseteq, \sqsupseteq]~as in Example~\ref{example:properties-example-1}. Finally, we look at \raacbrp[\supseteq_{pa}, \supseteq_{pro}, \sqsupseteq] where we combine both preferences over features with stages.

For all AA-CBR-based models, we \adamnew{use the empty set (and, where applicable, the empty sequence of stages) to characterise the default case}. The default outcome is set to progressive disease.

\subsection{Evaluation}


We evaluate the performance of the models using accuracy, precision, recall and F1-score, with progressive disease as the `positive' class~\cite{evaluation-metrics}.


Table~\ref{tab:results-static} shows results for the first experiment. We observe that \raacbrp[\supseteq_{pa}, \supseteq_{pro}]~outperforms the other models in terms of accuracy, recall, and F1 score. The decision tree and AA-CBR$_{\supseteq}$ perform similarly, but both are outperformed by the preference-based \raacbrp[\supseteq_{pa}, \supseteq_{pro}]~model. The introduction of preferences, in this instance, leads to improved classification performance but this cannot be done naively as with AA-CBR$_{\succcurlyeq_{lex}}$, which has the worst performance by almost all metrics. Clearly, our novel implementation that introduces preferences with \aacbrp~leads to an empirically better classifier for this task.

Table~\ref{tab:results-dynamic} shows the results for the second experiment. The decision-tree and k-nearest neighbor models are outperformed by the \aacbrp~models. Notably, we see that \raacbrp[\supseteq, \sqsupseteq]  has an increase in performance compared to AA-CBR with Stages, highlighting how forcing the model to conform to Theorem~\ref{theorem:preferred-cases} leads to more desirable empirical results. Furthermore, adding preferences over features with \raacbrp[\supseteq_{pa}, \supseteq_{pro}, \sqsupseteq] results in another increase in performance, again showcasing how preferences improve model performance.   

Overall, we find that \aacbrp~is not only successful at this classification task but can effectively leverage preferences to increase model performance. Moreover, our approach is flexible and generalised, so it can support varying representations, such as through the introduction of stages with preferences which improves performance further compared to AA-CBR. \adamnew{This comes at a cost of increasing time complexity by a factor of $m$, where $m$ is the number of partial orders}\footnote{\adamnew{Complexity results can be found in Appendix~\ref{section:complexity}.}}. 

\begin{table}
\centering
\setlength{\tabcolsep}{0.45\tabcolsep}
\begin{tabular}{lrrrr}
\toprule
Model  & Accuracy  & Precision & Recall & $F_{1}$  \\
\midrule

Decision Tree               & 0.72  & 0.83  & 0.63  & 0.72    \\
k-Nearest Neighbor          & 0.68  & \textbf{0.88}  & 0.50  & 0.64    \\

\midrule
AA-CBR$_{\supseteq}$           & 0.72  & 0.82  & 0.64  & 0.72    \\
AA-CBR$_{\succcurlyeq_{lex}}$  & 0.60  & 0.67  & 0.57  & 0.62    \\
\midrule
\raacbrp[\supseteq_{pa}, \supseteq_{pro}]   & \textbf{0.78}  & 0.81  & \textbf{0.79}  & \textbf{0.80}    \\
\bottomrule
\end{tabular}
\caption{Results of the first experiment where we utilise solely PA and PRO features to predict progressive disease in patients.}
\label{tab:results-static}
\end{table}

\begin{table}
\centering
\setlength{\tabcolsep}{0.45\tabcolsep}
\begin{tabular}{lrrrr}
\toprule
Model  & Accuracy  & Precision & Recall & $F_{1}$  \\
\midrule
Decision Tree               & 0.75  & \textbf{0.94}  & 0.58  & 0.72    \\
k-Nearest Neighbor          & 0.70  & 0.93  & 0.50  & 0.65    \\
\midrule
AA-CBR with Stages          & 0.74  & 0.86  & 0.64  & 0.73    \\
\midrule
\raacbrp[\supseteq, \sqsupseteq]   & 0.76  & 0.86  & 0.68  & 0.76    \\
\raacbrp[\supseteq_{pa}, \supseteq_{pro}, \sqsupseteq]   & \textbf{0.80}  & 0.88  & \textbf{0.75}  & \textbf{0.81}   \\
\bottomrule
\end{tabular}
\caption{Results of the second experiment where we utilise PA and PRO features in combination with a proxy measurement of previous disease progression to predict progressive disease in patients.}
\label{tab:results-dynamic}
\end{table}

\section{Related Work and Discussion}
\label{section:related-works}

Preferences over arguments have been shown to resolve situations in which the direction of an attack is unclear. Approaches often introduce an ordering specifying preferences over arguments,  as in~\cite{preference-argumentation-semantics,
preference-af-by-attacks}. Argumentation frameworks derived from AA-CBR with a single order can be viewed similarly, where the order defines preferences over the casebase. However, as shown in Example \ref{example:aa-cbr-motivating-1}, this alone is insufficient when introducing local preferences over constituents of an argument for classification. 
Existing approaches for accommodating preferences over constituents of arguments have also been proposed, e.g. as in \cite{ASPIC+,aba+,Gorgias,DeLP} for various structured argumentation frameworks. \adamnew{It is an open question as to whether these methods generalise or approximate our approach.} As future work, it would be interesting to study whether any of these approaches correspond to our method.

Our empirical evaluation shows the usefulness of accommodating domain expertise in the form of preferences 
within the medical setting.
It is widely acknowledged that healthcare providers have domain expertise that can influence the outcome of explainable AI systems~\cite{xai-in-healthcare}. Moreover, personalised healthcare and increased patient engagement can affect healthcare outcomes~\cite{patient-preferences}. 
Argumentation with preferences has already been shown to be effective for personalised healthcare. Specifically, in~\cite{aba+-patients}, the assumption-based argumentation model ABA$^{+}$G is used to resolve patient preferences with clinical guidelines. In that approach, preferences are defined over both assumptions, which dictate the direction of attacks and goals, in turn defining which extension(s) to accept. Also, the Gorgias system~\cite{Gorgias} applies argumentation with preferences to multiple healthcare scenarios, by allowing experts to express preferences over outcomes of scenarios which lead to preference hierarchies then translated into arguments about outcomes. Like with \aacbrp,~both these approaches use preferences defined over constituents of arguments. However, these approaches derive rules from a knowledge base, e.g. clinical guidelines, rather than by case-based reasoning.

\aacbrp\ is also related to 
case-based reasoning with precedential constraints, which has been extensively researched within the legal domain~\cite{rules-and-reasons-precedent,precedential-constraint-factors}. Dimension-based precedential constraints~\cite{precedential-constraint-dimensions,precedential-constraint-dimensions-2} represent an approach in which a set of values sampled from varying dimensions defines a case. Each dimension is equipped with a partial order in which values at either end of the partial order agree with opposing outcomes. In a classification task~\cite{precedent-based-incomplete-cases}, past cases constrain dimensions of a new case to an outcome consistent with the casebase. However, an outcome may not be predicted if the casebase cannot constrain a new case, and its outcome will remain undecided. 
 Previous work has shown how to enforce factor-based precedential constraints on $\text{AA-CBR}_\supseteq$ \cite{monotonicity-and-noise-tolerance}, deciding non-constrained cases,
 but it does not incorporate a notion of preferences between the factors, which is captured by the preferences in \aacbrp.
 Indeed, each preorder in $\pseq$~can be considered a generalised form of dimension in the precedential constraint sense. By introducing a lexicographic ordering over these `dimensions', we enforce that an outcome must be predicted, and preferred cases constrain new cases. Moreover, our approach does not enforce that either end of a partial order must agree with an outcome. This is suitable in a healthcare setting where such a restriction should not always apply; for example, sustained blood pressure that is either too high or too low could both have negative health consequences.

\section{Conclusion}
\label{section:conclusion}



In this paper, we introduced \aacbrp~as a novel method of including preferences with an AA-CBR-based approach. We have shown how preferences allow domain-specific knowledge or individual choices to guide the model and that existing approaches to AA-CBR cannot support this. Moreover, we prove that the desirable property of predictions abiding by the nearest cases, hold for \raacbrp~and present a stronger condition for preferred cases. Given this and that individual attacks always occur by the most preferred order, we conclude that the model inherently abides by the preferences by construction. We then show that the addition of preferences and the enforcement of the nearest and preferred cases conditions can empirically lead to increased performance in a medical classification task, outperforming that of other interpretable baseline models.

We have not explored methods of automatically deriving a preference ordering that directly optimises the model's performance. Methods such as the DEAr Methodology~\cite{DEAr} could be applied to utilise \raacbrp~to find characterisations and preference orderings that lead to the most optimal performance possible whilst still allowing for flexibility in user-defined preferences. Furthermore, we leave for future work the exploration of \raacbrp~variants, such as a cumulative model akin to cAA-CBR of~\cite{monotonicity-and-noise-tolerance}. 



\section*{Acknowledgements}



Adam Gould is supported by UK Research and Innovation [UKRI Centre for Doctoral Training in AI for Healthcare grant number EP/S023283/1]. 
Francesca Toni and Guilherme Paulino-Passos were partially funded by the 
ERC under
the EU’s Horizon 2020 research and innovation programme (grant agreement 
No. 101020934). Toni was also partially funded 
by 
J.P. Morgan and  the
Royal Academy of Engineering, UK, under the Research Chairs and Senior Research Fellowships scheme. 
Seema Dadhania is supported by the Imperial College CRUK Centre Clinical Research Fellowship A27434.
Matthew Williams is supported by the NIHR Imperial Biomedical Research Centre (BRC) and by ICHT NHS Trust.

We would like to thank the Computational Logic and Argumentation (CLArg) group and the Computational Oncology group, both at Imperial College London, for their feedback and support.
 



\bibliographystyle{kr}
\bibliography{references}
\clearpage

\appendix

\section{Theorem Proofs}
\label{section:proofs}

\subsection{Proof of Theorem \ref{theorem:preferred-cases}}



This proof generalises that of~\cite{monotonicity-and-noise-tolerance-appendix}.

\begin{proof} 

We begin by establishing that for a coherent $D$, and when every preferred case to $N$ is in the form of $(\xalpha, y)$, each argument in $G_i$ is either labelled as $N$ or corresponds to a case $(\xbeta, y)$, where $\mathbb{G}$ is the union of all $G_i$ sets. In simpler terms, every case within the subsets of the grounded set is either labelled as $N$ or shares the same outcome as the preferred cases. We prove this by induction on $i$:
    \begin{enumerate}
        \item For the base case, we must show this property holds for $G_{0}$ which contains only unattacked arguments. $N$ is always unattacked and so is an element of $G_{0}$. Other unattacked cases may also be in $G_{0}$ so we let $\beta = \casebeta \in G_{0}$ be such a case. We must show that $y_{\beta} = y$. If $\beta$ is preferred for $N$ then $y_{\beta} = y$ as required. We now consider if $\beta$ is not preferred for $N$. If $\exists i, x_{N} \not \succcurlyeq_{i} \xbeta$, then, as we are dealing with regular \raacbrp[],~$N \nsim \beta$, leading to $N \newattacks \beta$ which cannot be as $\beta$ is unattacked. So we must have have that $x_{N} \bigsucccurlyeq{1}{n} \xbeta$. Assume towards contradiction that $y_{\beta} \neq o$.
        As $\beta$ is not a preferred case, there must exist a preferred case $\alpha = (\xalpha, y)$ such that $\xalpha \potentialattacks{i} \xbeta$ for some $i$. We know $\alpha$ must exist by definition of preferred cases
        and because $D$ is finite%
        .  By Definition \ref{def:specificity-attacks}, either $\alpha \attacks[i] \beta$ or there exists a case $\gamma = (\xgamma, y)$ that is a more concise attacker of $\beta$ and we have $\exists l \geq i, \gamma \attacks[l] \beta$. In either circumstance, this cannot be as $\beta$ must be unattacked. Thus, we have a contradiction, so our previous assumption, that $y_{\beta} \neq o$, does not hold and we have $y_{\beta} = y$ as required.



        \item In the inductive step, we assume the property holds for $G_{i}$ and show it holds for $G_{i+1}$. Let $\beta = \casebeta \in G_{i+1} \backslash G_{k}$ (if $\beta \in G_{i}$, the property holds by the induction hypothesis). So, by definition of the grounded extension, $\beta$ is defended by $G_{i}$, which is conflict-free so $N$ cannot attack $\beta$, thus we have $x_{N} \bigsucccurlyeq{1}{n} \xbeta$. Again, if $\beta$ is a preferred case to $N$, then $y_{\beta} = y$ as required. We now consider when $\beta$ is not a preferred case. Assume towards contradiction that $y_{\beta} \neq o$. As $\beta$ is not a preferred case, there must exist a preferred case $\alpha = (\xalpha, y)$ such that $\xalpha \potentialattacks{i} \xbeta$ for some $i$.  As before, either $\alpha \attacks[i] \beta$ or there exists a case $\gamma = (\xgamma, y)$ that is a more concise attacker of $\beta$, and we have $\exists l \geq i, \gamma \attacks[l] \beta$. Let $\eta$ be the attacker of $\beta$. As $G_{i}$ defends $\beta$, there must exist a $\theta \in G_{i}$ such that $\theta \attacks \eta$. By the inductive hypothesis, $\theta$ is either $N$ or $\theta = (x_{\theta}, y)$. $\theta$ cannot be $N$ as for $\eta$ to be equal to $\alpha$ or a more concise case than $\alpha$, we must have $\xalpha \bigsucccurlyeq{1}{n} x_{\eta}$ and as $\alpha$ is a preferred case to $N$, we have $x_{N} \bigsucccurlyeq{1}{n} \xalpha$. So $x_{N} \bigsucccurlyeq{1}{n} x_{\eta}$, which means $\eta$ is not irrelevant to $N$ and, thus, cannot be attacked by $N$. So we have instead $\theta = (x_{\theta}, y)$. But in this case, $\theta$ cannot possibly attack $\eta$ in defence of $\beta$ because $y_{\theta} = y_{\eta} = y$. We, therefore, have a contradiction, so $y_{\beta} = y$ as required.

    \end{enumerate}

    We can thus conclude that every argument in the grounded set, $\mathbb{G}$, is either $N$ or of the form $(\xbeta, y)$.

    As a consequence, if $y = \bar{\delta}$, (that is all preferred cases agree on the non-default outcome), then $\casedefault \not \in \mathbb{G}$ because as we have just shown, all arguments in $\mathbb{G}$ excluding $N$ must have the outcome $o$. As a result, the predicted outcome for $N$ must be $o$ as required by the theorem.

    If $y = \delta$, then we must show that $\casedefault \in \mathbb{G}$. If $\casedefault$ is unattacked, the default case is in the grounded extension as required. Consider instead if $\casedefault$ is attacked by a case $\beta$.
    As we are dealing with regular \raacbrp[], $\casedefault$ can never be irrelevant to $N$ and so $\beta$ cannot be $N$.
    Thus, $\beta$ is of the form $(\xbeta, \bar{\delta})$ and is therefore not in $\mathbb{G}$ because, as have shown, all arguments in $\mathbb{G}$ must have outcome $o$. We know that $D$ is coherent, so the AF is acyclic by Proposition~\ref{prop:acyclic} and $\mathbb{G}$ is a stable extension, meaning every argument not in $\mathbb{G}$ is attacked by an argument in it. So there exists some argument in $\mathbb{G}$ that attacks $\beta$ and defends the default argument. By definition, $\mathbb{G}$ contains every argument it defends, so the default argument is also in $\mathbb{G}$. Therefore, the outcome predicted for $N$ will be $o$ as required by the theorem.

\end{proof}

\subsection*{Proof of Theorem \ref{theorem:nearest-neighbour}}





\begin{proof}

    We first show that given the finite set, $\Gamma$, of all cases nearest to $N$ in $D$, there is always at least one case that is preferred for $N$. Assume towards contradiction that no cases in $\Gamma$ are preferred for $N$. So for a case, $\casealpha \in \Gamma$ there exists another case, $\casebeta \in \Gamma$ such that $\exists j,$ $\xalpha \bigeq{1}{j - 1} \xbeta$ and $\xbeta \succ_{j} \xalpha$ by definition of preferred cases. We can say that $\alpha$ is \emph{less preferred} than $\beta$ or $\alpha \ll \beta$. We show this relation is transitive: if we assume $\alpha \ll \beta$ and $\beta \ll \gamma$, then we know that $\xalpha \bigeq{1}{j - 1} \xbeta$ and $\xbeta \succ_{j} \xalpha$ for some $j$ and $\xbeta \bigeq{1}{k - 1} \xgamma$ and $\xgamma \succ_{k} \xbeta$ for some $k$. As a result, we have that $\xalpha \bigeq{1}{\min(j, k) - 1} \xgamma$ and $\xalpha \succ_{\min(j, k)} \xgamma$ which gives us that $\alpha \ll \gamma$.
    

    So for cases $\alpha, \beta \in \Gamma$, if we have that $\alpha \ll \beta$ and we know that $\beta$ is not a preferred case, then there must exist another case, $\gamma \in \Gamma$, in which $\beta \ll \gamma$. However, by our assumption, $\gamma$ is also not a preferred case. As $\Gamma$ is finite, there must be a sequence in which $\alpha \ll \beta \ll \gamma \ll \ldots \ll \gamma$. But this cannot be as it implies by transitivity of $\ll$ that $\exists j, \xgamma \succ_{j} \xgamma$, which is a contradiction. Therefore, $\Gamma$ must contain at least a single preferred case.

   If all nearest cases agree on an outcome, then all the preferred cases, of which we have shown that at least one must exist, will agree on an outcome, and so by Theorem \ref{theorem:preferred-cases}, we have proved this theorem as required. 

\end{proof}

\subsection{Proof of Proposition \ref{prop:acyclic}}


\begin{proof}
First, as $D$ is coherent, the set of incoherent attacks $\incoherentattacks$ will be empty, so there will not be any symmetric attacks. Secondly, as $N$ cannot be attacked, it cannot be part of cycles in the graph. As $\newattacks$ only originates from $N$, we do not need to consider these.  

We now consider only \mbox{$\attacks = (\bigcup_{i=1}^{n} \attacks[i])$}, where $n$ corresponds to the size of $P_{\succcurlyeq}$. For any attack $\casealpha \attacks \casebeta$, we know that either $(\xalpha \succ_{1} \xbeta)$ or $(\exists i > 1, \xalpha =_{1} \xbeta$ and $\xalpha \succ_i \xbeta)$. This means that across a path of attacks, $(\xalpha, y_{\alpha}) \attacks \ldots \attacks (\xbeta, y_{\beta}) \attacks \ldots \attacks (\xgamma, y_{\gamma})$, we have that $\xalpha \succcurlyeq_{1} \xgamma$. If at any point in the path of attacks, we have $\xalpha \succ_{1} \xbeta$ then we must have $\xalpha \succ_{1} \xgamma$ because a case in a path of attacks can never get more specific on the first order. If, say, all cases in the path were all equal on the first order, that is $\xalpha =_{1} \xbeta =_{1} \ldots =_{1} \xgamma$, then attacks in the path will never need to consider the first order by the lexicographic ordering of the specificity relations and so we can apply the previous reasoning for $\succcurlyeq_{2}$. That is, along the path of attacks, we must have $\xalpha \succcurlyeq_{2} \xgamma$ and if $\xalpha \succ_{2} \xbeta$ then $\xalpha \succ_{2} \xgamma$. We can extend this reasoning arbitrarily until the last order. 

If we assume towards a contradiction that we have a cycle, we will have a path such that $(x_{1}, y_{1}) \attacks (x_{2}, y_{2}) \attacks \ldots \attacks (x_{q}, y_{q}) \attacks (x_{1}, y_{1})$. We know that the first and last cases in this path are equal in the first order, and as a path of attacks can never get more specific in the first order, all cases in this path must be equal in the first order, that is $x_{1} =_{1} x_{2} =_{1} \ldots =_{1} x_{q} =_{1} x_{1}$. As we apply lexicographic ordering and all the cases are equal on the first order, this path of attacks only ever considers orders from $\succcurlyeq_{2}$ onwards. But again, we know that the first and last cases are equal in the second order, so we must have $x_{1} =_{2} x_{2} =_{2} \ldots =_{2} x_{q} =_{2} x_{1}$. We apply this reasoning across all orders until we eventually get that the only possible attacks that can occur are on the last order, $\succcurlyeq_{n}$. But then we would have $x_{1} \succ_{n} x_{1}$ which is a contradiction.

\end{proof}

\section{Deriving Existing Formulations from \aacbrp}
\label{section:derivations}
\subsection{Deriving AA-CBR from \aacbrp}



To capture Definition \ref{def:aa-cbr_geq} in \aacbrp, we must show that we can precisely derive $Args$ and~$\attacks$ when instantiating with $\pseq = \langle \succcurlyeq \rangle$ where $\succcurlyeq$ is an arbitrary partial order defined over the characterisations of the casebase. By Definition \ref{def:aa-cbr-p}, we have: 

\begin{itemize}
    \item  $Args = D \cup \{\casedefault\} \cup \{N\}$, 
    \item \mbox{$\attacks =  \attacks[1] \cup \incoherentattacks \cup \newattacks$}. 
\end{itemize}

$Args$ is as required. We will now show that $\attacks$ corresponds to that of Definition~\ref{def:aa-cbr_geq}. By Definition \ref{def:specificity-attacks} we have $\alpha \attacks[1] \beta$ iff

    \begin{enumerate}[i]
        \item $\alpha \potentialattacks{1} \beta$ and
        
        \item $\nexists \gamma = \casegamma \in D \cup \{ \casedefault \}$ with $\xalpha \succcurlyeq_{1} \xgamma$ and
              \begin{enumerate}
                  \item either $\gamma \potentialattacks{1} \beta$ and $\exists l \geq 1,$ $\xalpha \succ_{l} \xgamma$, 
                  
                  \item or $\gamma \not \potentialattacks{1} \beta$ and $\exists l > 1,$ $\gamma \potentialattacks{l} \beta$. 
                  
              \end{enumerate}
    \end{enumerate}

We know that condition ii (b) cannot hold as there is only one order. By this same reasoning, we also know that in condition ii (a), we can only let $l = 1$.  Furthermore, by utilising Definition~\ref{def:potential-attacks} we get $\alpha \attacks[1] \beta$ iff

\begin{enumerate}[i]
    \item $y_{\alpha} \neq y_{\beta}$, and
    \item $\xalpha \succ_1 \xbeta$, and
        
    \item $\nexists \gamma = \casegamma \in D \cup \{ \casedefault \}$ and $\xgamma \succ_{1} \xbeta$ and $\xalpha \succ_{1} \xgamma$, 
\end{enumerate}

Which is logically equivalent to the second bullet point, points 1. and 2. of Definition~\ref{def:aa-cbr_geq}. 

By Definition \ref{incoherent-attacks}, we have $\casealpha \incoherentattacks \casebeta$ iff \mbox{$y_{\alpha} \neq y_{\beta}$}, and $\xalpha =_{1} \xbeta$. This is logically equivalent to bullet 2, point 1 and point 3. of Definition~\ref{def:aa-cbr_geq}.

Finally, by Definition~\ref{new-case-attacks}, we have $N \newattacks \casealpha$ iff $N \nsim \casealpha$, which is the same as the final bullet of Definition~\ref{def:aa-cbr_geq}.

Putting this all together gives:

    \begin{itemize}
        \item $Args = D \cup \{\casedefault\} \cup \{N\}$
        \item for $\casealpha, \casebeta \in D \cup \{(\x{\delta}, y_\delta)\}$, it holds that $\casealpha \rightsquigarrow \casebeta$ iff
              \begin{enumerate}
                  \item $y_{\alpha} \not = y_{\beta}$, and
                  \item either $\x{\alpha} \succ \xbeta$ and $\not \exists \casegamma \in D \cup \{(\x{\delta}, y_{\delta})\}$ with $\xalpha \succ \xgamma \succ \xbeta$  \hfill
                  \item or $\xalpha = \xbeta$
              \end{enumerate}
        \item for $\casealpha \in D \cup \{(\x{\delta}, y_{\delta})\}$, it holds that $N \rightsquigarrow \casealpha$ iff $N \nsim \casealpha.$
    \end{itemize}

which is precisely Definition~\ref{def:aa-cbr_geq}.

\subsection{Deriving a variant of AA-CBR With Stages from \raacbrp}

We aim to capture Definition \ref{def:aa-cbr-with-stages} in \raacbrp~but with condition \textbf{2. (b) i} altered to $F_{\alpha} \supset F_{\gamma} \supset F_{\beta}$ and $S_{\alpha} \sqsupseteq S_{\gamma}$. Moreover, Definition \ref{def:aa-cbr-with-stages} was defined only for a coherent casebase and does not include symmetric attacks, we will include these in our derivation. We will refer to this as capturing a `variant of Definition\ref{def:aa-cbr-with-stages}'.
Note that we use a regular variant here because AA-CBR with Stages is defined utilising the default case containing the smallest elements by the~$\supseteq$ and~$\sqsupseteq$ relations and $\nsim$ is defined in terms of these orders akin to Definition~\ref{def:regular-aa-cbr-p}.

We must show that we can derive $Args$ and~$\attacks$ from Definition \ref{def:aa-cbr-p}. We let $\pseq = \langle \supseteq, \sqsupseteq \rangle$, and $x_{\delta} = (\emptyset, \langle \rangle)$. By Definition \ref{def:aa-cbr-p}, we have that $Args = D \cup \{\casedefault\} \cup \{N\}$ as required. 

Again by Definition~\ref{def:aa-cbr-p}, we have $\attacks =  \attacks[1] \cup \attacks[2] \cup \incoherentattacks \cup \newattacks$.  

By Definition \ref{def:specificity-attacks}, we have $\alpha \attacks[1] \beta$ iff
    \begin{enumerate}[i]
        \item $\alpha \potentialattacks{1} \beta$ and
        \item $\nexists \gamma = \casegamma \in D \cup \{ \casedefault \}$ with $\xalpha \succcurlyeq_{1} \xgamma$ and $\xalpha \succcurlyeq_{2} \xgamma$ and
              \begin{enumerate}
              
                  \item either $\gamma \potentialattacks{1} \beta$ and $\exists l \geq 1,$ $\xalpha \succ_{l} \xgamma$, 
                  
                  \item or $\gamma \not \potentialattacks{1} \beta$ and $\exists l > 1,$ $\gamma \potentialattacks{l} \beta$. 
                  
              \end{enumerate}
    \end{enumerate}

As there are two orders, we know that in ii (a), we can set $l$ to either 1 or 2 and in ii (b) $l$ can only be set to 2. Substituting this we get $\alpha \attacks[1] \beta$ iff

\begin{enumerate}[i]
    \item $\alpha \potentialattacks{1} \beta$ and
    \item $\nexists \gamma = \casegamma \in D \cup \{ \casedefault \}$ with $\xalpha \succcurlyeq_{1} \xgamma$ and $\xalpha \succcurlyeq_{2} \xgamma$ and
          \begin{enumerate}
          
              \item either $\gamma \potentialattacks{1} \beta$ and ($\xalpha \succ_{1} \xgamma$ or  $\xalpha \succ_{2} \xgamma$),
              
              \item or $\gamma \not \potentialattacks{1} \beta$ and $\gamma \potentialattacks{2} \beta$. 
              
          \end{enumerate}
\end{enumerate}

By applying Definition \ref{def:potential-attacks} it holds that $\alpha \attacks[1] \beta$ iff

\begin{enumerate}[i]
    \item $y_{\alpha} \neq y_{\beta}$, and
    \item $\xalpha \succ_1 \xbeta$, and
    \item $\nexists \gamma = \casegamma \in D \cup \{ \casedefault \}$ with $\xalpha \succcurlyeq_{1} \xgamma$ and $\xalpha \succcurlyeq_{2} \xgamma$ and
          \begin{enumerate}
          
              \item either $\xgamma \succ_1 \xbeta$ and ($\xalpha \succ_{1} \xgamma$ or  $\xalpha \succ_{2} \xgamma$),
              
              \item or   $\xgamma \succ_2 \xbeta$ and $\xgamma =_{1} \xbeta$.
              
          \end{enumerate}
\end{enumerate}

Note that we use the fact that $\gamma \potentialattacks{2} \beta$ implies $\gamma \not \potentialattacks{1} \beta$ to remove the latter term. We can rearrange this to get $\alpha \attacks[1] \beta$ iff

\begin{enumerate}[i]

    \item $y_{\alpha} \neq y_{\beta}$, and
    \item $\xalpha \succ_{1} \xbeta$, and
    \item $\not \exists \dynamiccasegamma  \in D$ with
    \begin{enumerate}[i]
      \item either  $\xalpha \succ_{1} \xgamma \succ_{1} \xbeta$ and $\xalpha \succcurlyeq_{2} \xgamma$,
      \item or $\xgamma =_{1} \xalpha$ and $\xalpha \succ_{2} \xgamma$
      \item or $\xbeta =_{1} \xgamma$ and $\xalpha \succcurlyeq_{2} \xgamma \succ_{2} \xbeta$
    \end{enumerate}
\end{enumerate}

which corresponds to bullet 3, point 2, (a) and (b) of the modified variant of Definition \ref{def:aa-cbr-with-stages} as required. 

Again by Definition \ref{def:specificity-attacks}, we have $\alpha \attacks[2] \beta$ iff

\begin{enumerate}[i]
    \item $\alpha \potentialattacks{2} \beta$ and
    \item $\nexists \gamma = \casegamma \in D \cup \{ \casedefault \}$ with $\xalpha \succcurlyeq_{1} \xgamma$ and $\xalpha \succcurlyeq_{2} \xgamma$ and
          \begin{enumerate}
          
              \item either $\gamma \potentialattacks{2} \beta$ and $\exists l \geq 2,$ $\xalpha \succ_{l} \xgamma$, 
              
              \item or $\gamma \not \potentialattacks{2} \beta$ and $\exists l > 2,$ $\gamma \potentialattacks{l} \beta$. 
              
          \end{enumerate}
\end{enumerate}

As the length of $\pseq$ is 2, we know that in ii (a), $l$ must be set to 2 and ii (b) does not hold. We therefore get $\alpha \attacks[2] \beta$ iff

\begin{enumerate}[i]
    \item $\alpha \potentialattacks{2} \beta$ and
    \item $\nexists \gamma = \casegamma \in D \cup \{ \casedefault \}$ with $\xalpha \succcurlyeq_{1} \xgamma$ and $\xalpha \succcurlyeq_{2} \xgamma$ and
          \begin{enumerate}
              \item $\gamma \potentialattacks{2} \beta$ and $\xalpha \succ_{2} \xgamma$. 
          \end{enumerate}
\end{enumerate}

\noindent
Applying Definition \ref{def:potential-attacks}, it holds that $\alpha \attacks[2] \beta$ iff

\begin{enumerate}[i]
    \item $y_{\alpha} \neq y_{\beta}$, and
    \item $\xalpha \succ_2 \xbeta$, and $\xalpha =_{1} \xbeta$.
    \item $\nexists \gamma = \casegamma \in D \cup \{ \casedefault \}$ with $\xalpha \succcurlyeq_{1} \xgamma$ and $\xalpha \succcurlyeq_{2} \xgamma$ and
          \begin{enumerate}
              \item $\xgamma =_{1}  \xbeta$ and $\xgamma \succ_{2}  \xbeta$ and $\xalpha \succ_{2} \xgamma$. 
          \end{enumerate}
\end{enumerate}

\noindent
This corresponds to bullet 3, point 3 of the modified variant of Definition \ref{def:aa-cbr-with-stages} as required. 

As we are using the regular variant, $\raacbrp$, we have by Definitions~\ref{new-case-attacks} and \ref{def:regular-aa-cbr-p}: $N \newattacks \casealpha$ iff $\exists i, x_{N} \not \succcurlyeq_{i} \xalpha$. This is equivalent to: $N \newattacks \casealpha$ iff $x_{N} \not \succcurlyeq_{1} \xalpha$ or $x_{N} \not \succcurlyeq_{2} \xalpha$ as required. This corresponds to the final bullet of Definition~\ref{def:aa-cbr-with-stages}.

For symmetric attacks, by Definition~\ref{incoherent-attacks}, we have that $\casealpha \incoherentattacks \casebeta$ iff $\xalpha =_{1} \xbeta$ and $\xalpha =_{2} \xbeta$ and $y_{\alpha} \neq y_{\beta}$ as required.

\section{Complexity Results}
\label{section:complexity}
\adamnew{The time complexity of \aacbrp~can be divided into three main parts: AF construction with the casebase, identification of new case attacks, and computing the grounded extension. The construction of the AF using the casebase is a one-time cost that can be amortised over the number of times that a prediction is made for a new case. Assuming that each partial order takes $O(1)$ to compute, the one-time cost of AF construction has a worst-case time complexity of $O(n^{3}m)$ where $n$ is the size of the casebase, and $m$ is the number of partial orders in $\pseq$. This is compared to AA-CBR, which has a worst-case time complexity of $O(n^{3})$. For regular $\raacbrp$, the time complexity of finding new case attacks is $O(nm)$, compared to $O(n)$ for AA-CBR. The computation of the grounded extension is the same for both AA-CBR and $\aacbrp$ and is known to be in P \cite{aa-complexity}. The increase in time complexity when adding preferences to AA-CBR is marginal given that it is expected that $m << n$.}

\section{Experiment Methodology}
\label{section:experiment-methods}
 \subsection{Data}


Data from 78 patients was provided by the BrainWear study~\cite{brainwear-paper}. \adamnew{All patient data was collected with informed, written consent. The original study was approved by the South West-Cornwall \& Plymouth Research Ethics Committee (18/SW/0136). The use of real-world healthcare data within this context showcases how \aacbrp~can be used to create explainable AI models that are intrinsically interpretable and thus suitable to be applied within a high-stakes decision-making scenario.} 
17 Patients without a high-grade glioma (the disease of focus) were excluded, followed by 2 patients with insufficient Physical Activity data quality, 21 patients who did not complete at least two PRO questionnaires and 5 who lacked data from a least one scan. This left 31 eligible patients. 

We extracted 110 data points from the 31 patients with applicable data. Each data point can be represented by a binary set of features extracted from a questionnaire and the surrounding physical activity data and labelled by the outcome of the MRI scan following the questionnaire or survival status at the end of the study. Additionally, each data point can be assigned a proxy time measure of the patient's cancer journey.  
 
 \subsubsection{Patient Reported Outcome (PRO) Questionnaires}

During the study, patients periodically completed questionnaires reporting their symptoms, outlooks, ability to function, and overall quality of life. We focus on the EORTC QLQ-C30 questionnaire and brain tumour-specific BN20 module~\cite{eortc-qlq-c30}. Each scale is scored from 1-100~\cite{eortc-scoring-manual}. We utilised the following five scales: fatigue $(fa)$, global health QoL $(ql)$, physical functioning $(pf)$, future uncertainty $(fu)$ and motor dysfunction $(md)$ in accordance with~\cite{brainwear-paper}. For each scale, a binary feature is extracted, denoting an observed change of at least 50\% compared to a baseline questionnaire completed by each patient. We take an increase and a decrease of 50\% as distinct features. We, therefore, have the following PRO features: $\mathbb{F}_{pro} = \{faI, faD, qlI, qlD, pfI, pfD, fuI, fuD, mdI, mdD$\} where the suffixes `I' and `D' represent an increase and decrease of the scale respectively.

\subsubsection{Physical Activity Data}

 Every patient was given an Axivity AX3 wrist-worn accelerometer, which measures acceleration in milligravity (mg) units in the x, y, and z directions at 100 Hz. Data was processed with the UK Biobank accelerometer analysis pipeline, which provided an average acceleration value for each 30-second epoch and a classification of the likely physical behaviour of the patient at that time. These behaviours are sleep $(sl)$, sedentary behaviour $(sb)$, conducting light tasks $(lt)$, moderate activity $(ma)$ and walking $(wa)$~\cite{biobank-processing-1,biobank-processing-2,biobank-processing-3,biobank-processing-4}. A data point could be extracted if there exists physical activity data covering 50\% of an 8-week period centred on a completed questionnaire. For each patient, the time spent conducting each behaviour was averaged across the eight weeks. A binary feature denotes if the time spent doing each behaviour changed by at least 50\%, compared to a baseline measurement, taken as an average of the PA for the first two weeks each patient was in the study. We, therefore, have the following PA features $\mathbb{F}_{pa} = \{slI, slD, sbI, sbD, ltI, ltD, maI, maD, waI, waD \}$.

\subsubsection{Patient Outcomes}

Patients had recurring MRI scans over the course of the study. The raw imaging data was not provided, but a clinician labelled each image, indicating that the patient's disease was either progressive or stable. Each data point was labelled utilising the first MRI scan following the completed questionnaire, with an outcome of 1 representing progressive disease (pd) and 0 representing stable disease (sd). If there was no MRI scan following a questionnaire, we took the outcome by the survival status of the patient at the time of data collation, with 1 indicating that the patient had died at this time and 0 indicating otherwise. Thus, labelling mirrors the concept of progression-free survival.

Furthermore, we can consider the number of times a patient has previously had an MRI showcasing progressive disease at the time of the completed questionnaire as a proxy measure for the progression of their cancer. We could, therefore, define stages (as in the notion of AA-CBR with Stages), $\mathbb{S}_{pd} = \langle pd_{0}, pd_{1}, pd_{2}, pd_{3} \rangle$.

\subsubsection{Data Points}
Of the 110 data points, 60 have been used as training data (thus making up the casebase for the AA-CBR-based models), and 50 are held out for testing.

\subsection{Baseline Models}
The decision tree (gini criteria) was selected due to providing tree-like explanations which can also be provided for AA-CBR~\cite{arbitrated-argumentative-dispute}. The kNN (with k=3 and utilising hamming distance) was selected because this model also utilises case-based reasoning. For these baseline models, a binary feature vector is constructed, where for each value in $\mathbb{F}_{pa} \cup \mathbb{F}_{pro}$, a 1 represents that the feature is present and a 0 otherwise. Stages were one-hot encoded and appended to the feature vector for experiments that utilised them.  



\end{document}